\documentclass[runningheads]{llncs}

\usepackage{eccv}

\usepackage{eccvabbrv}

\usepackage{graphicx}
\usepackage{booktabs}

\usepackage[accsupp]{axessibility}  

\usepackage{hyperref}

\usepackage{orcidlink}

\usepackage[utf8]{inputenc} 
\usepackage[T1]{fontenc}    
\usepackage{hyperref}       
\usepackage{url}            
\usepackage{booktabs}       
\usepackage{amsfonts}       
\usepackage{nicefrac}       
\usepackage{microtype}      
\usepackage{xcolor}         

\usepackage{amsmath}
\usepackage{amssymb} 
\usepackage{mathtools}
\usepackage[dvipsnames]{xcolor} 
\usepackage{colortbl}
\usepackage{multirow}
\usepackage{nicematrix}
\usepackage{subcaption}
\usepackage[thicklines]{cancel}
\usepackage{color}
\usepackage{enumitem}

\newcommand{\beginsupplement}{
    \appendix
	\setcounter{table}{0}
	\renewcommand{\thetable}{A\arabic{table}}%
	\setcounter{figure}{0}
	\renewcommand{\thefigure}{A\arabic{figure}}%
	\setcounter{equation}{0}
	\renewcommand{\theequation}{A\arabic{equation}}
}

\DeclarePairedDelimiter{\brc}{(}{)}

\def\paperTitle{Step-by-Step Video-to-Audio Synthesis \\ via Negative Audio Guidance}

\newcommand*\xcond{x^{(1)}}
\newcommand*\xtgt{x^{(2)}}
\newcommand*\CE{\mathcal{E}}
\newcommand*\xcondc{\CE\brc*{x^{(1)}}}
\newcommand*\notxcondc{\bar{\CE}\brc*{x^{(1)}}}
\newcommand*\condc{\CE\brc*{\cdot}}
\newcommand*\notcondc{\bar{\CE}\brc*{\cdot}}

\newcommand*\PMfull{Negative Audio Guidance}
\newcommand*\PMabbr{NAG}

\newcommand*\fdvgg{FD\(_{\text{VGG}}\)}
\newcommand*\fdpanns{FD\(_{\text{PANNs}}\)}

\newcommand*\klpanns{KL\(_{\text{PANNs}}\)}
\newcommand*\klpasst{KL\(_{\text{PaSST}}\)}
\newcommand*\clapaa{CLAP A-A}
\newcommand*\clapta{CLAP T-A}

\definecolor{headerColor}{RGB}{224, 224, 224}
\definecolor{highlightColor}{RGB}{230, 244, 252}

\newcommand*{\deemph}[1]{\textcolor{gray}{#1}}
\newcommand\pmnum[1]{\scriptsize$\pm$#1}

\begin{document}

\title{\paperTitle{}} 

\titlerunning{Step-by-step V2A via Negative Audio Guidance}

\author{
Akio Hayakawa\inst{1}\orcidlink{0009-0007-5769-5711}
\and
Masato Ishii\inst{1}\orcidlink{0009-0005-6256-9264} 
\and
Takashi Shibuya\inst{1}\orcidlink{0000-0002-4277-0164}
\and
Yuki Mitsufuji\inst{1,2}\orcidlink{0000-0002-6806-6140}
}

\authorrunning{A.~Hayakawa et al.}

\institute{Sony AI, Tokyo, Japan
\and 
Sony Group Corporation, Tokyo, Japan
\email{\{akio.hayakawa,masato.a.ishii,takashi.tak.shibuya,yuhki.mitsufuji\}@sony.com}}

\maketitle

\begin{abstract}

We propose a step-by-step video-to-audio (V2A) generation method that provides finer control over the generation process and more realistic audio synthesis.
Inspired by traditional Foley workflows, our approach enables incremental generation of \emph{complementary} sounds, allowing users to author multiple sound events induced by a video.
To avoid the need for costly multi-reference video-audio datasets, each generation step is formulated as a negatively guided V2A process that discourages duplication of sounds already present in previously generated tracks.
The guidance model is trained by finetuning a pre-trained V2A model on audio pairs from non-overlapping segments of the same video, encouraging it to leverage acoustic context while remaining visually grounded, and enabling training with standard single-reference audiovisual datasets.
Objective and subjective evaluations demonstrate that our method enhances the separability of generated sounds at each step and improves the overall quality of the final composite audio, outperforming existing baselines.
Our project page is available at: \url{https://ahykw.github.io/sbsv2a/}.

\end{abstract}

\begin{figure}[t]
    \centering
    \includegraphics[width=\linewidth]{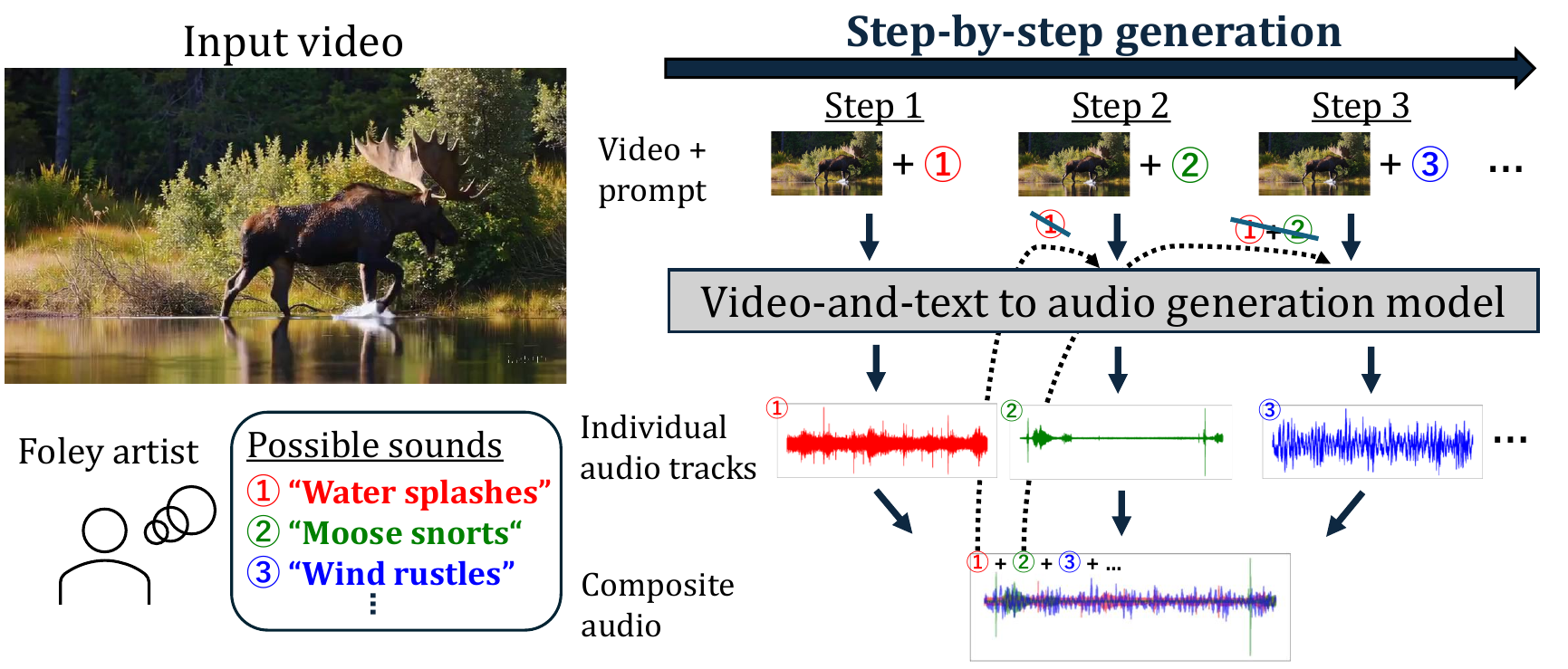}
    \caption{Step-by-step video-to-audio generation for compositional sound effect creation. Video often contains numerous audible events, and Foley artists synthesize composite audio by adding complementary audio components step-by-step. Supporting this step-by-step mechanism with a video-to-audio generation model offers greater control and efficiency in the sound creation process.}
    \label{fig:teaser}
\end{figure}

\section{Introduction}

Generating realistic audio signals that align seamlessly with given visual content is a process referred to as Foley~\cite{Foley} in film or game production.
In traditional workflows, Foley artists begin with field-recorded or library sounds and incrementally layer in missing elements (e.g., footsteps or fabric movements) to enhance audio realism.
While essential for high-quality audiovisual content, this workflow is labor-intensive and time-consuming because even short clips often contain numerous audible events.

Recent video-to-audio (V2A) models~\cite{viertola2025temporally,luo2023diff,wang2024frieren,wang2024v2a,liu2024tell,cheng2024taming,polyak2025moviegencastmedia} show promise for automating this workflow.
These models produce high-quality audio that semantically and temporally aligns with input videos.
However, most models generate an entire track in a single pass and do not offer a mechanism for incremental refinement (i.e., supplementing sounds missing in the generated results).
This non-interactive design poses a significant challenge: if the output is missing specific events, creators must regenerate the entire track.
Such inefficiencies limit the practical application of these models, particularly in collaborative workflows with human creators.

To resolve this issue, we argue that a step-by-step generation mechanism is crucial for practical V2A synthesis (Fig.~\ref{fig:teaser}). 
A model should generate not only a complete track aligned with the video, but also complementary audio that fills missing events without duplicating sounds already existing.\footnote{One might consider text-conditional V2A already provides sufficient control for the target audio event to be generated. However, existing text-conditional V2A models struggle to suppress the already generated sound, especially for prominent events in the video (e.g., in Fig.~\ref{fig:step-by-ste-visualize}, the moose's footstep sounds are produced in all tracks regardless of the input text prompts).}
This offers greater control and efficiency in the sound creation process, as in the traditional Foley workflow.

A critical challenge to achieve this step-by-step generation is the scarcity of datasets. 
A straightforward approach (\ie, training a conditional generation model that produces multiple plausible audio tracks per video) requires multi-reference video-audio pairs, which are difficult to obtain at scale. 
In this paper, we propose a guided generation method, \PMfull~(\PMabbr), that enables step-by-step V2A synthesis without requiring specialized multi-reference datasets.
Our key idea is to introduce an audio-conditioned branch trained on standard single-reference audiovisual datasets, and to use it \emph{negatively} at inference time to discourage duplication of previously generated sounds.
Concretely, we finetune a pre-trained V2A model using pairs of audio segments sampled from the same video, and during sampling, we apply \PMabbr\ to push the current generation away from the audio content already present in previously generated tracks.
By iterating this negatively guided generation, the model produces a set of complementary tracks that can be mixed into a composite audio.
Extensive experiments demonstrate that our method enables step-by-step completion of complementary sounds and enhances final audio quality while ensuring the separability of the generated audio at each step. In short, our main contributions are summarized as follows:
\begin{itemize}
    \setlength{\itemsep}{0.2em}
    \item We formulate \emph{step-by-step} V2A synthesis as generating multiple \emph{complementary} audio tracks for a single video, thereby enabling Foley-style interactive authoring.
    \item We propose \PMfull~(\PMabbr), which uses an audio-conditioned guidance branch \emph{negatively} during sampling to reduce duplication across sequentially generated tracks, while being trainable on standard single-reference datasets.
    \item We demonstrate improved track separability and composite audio quality compared to independent multi-prompt generation and text-based negative prompting.
\end{itemize}

\section{Related work}

\subsection{Video-to-audio synthesis}

The goal of video-to-audio synthesis is to generate an audio signal that aligns semantically and temporally with an input video. 
Early approaches used regression models~\cite{chen2020generating} and GANs~\cite{iashin2021taming}, while more recent ones have adopted autoregressive models~\cite{viertola2025temporally} and diffusion models~\cite{luo2023diff,wang2024frieren,wang2024v2a,liu2024tell,polyak2025moviegencastmedia,chen2024multifoley,cheng2024taming,Jun2025klingfoley} due to their high capability in generation tasks. 
However, these models typically accept only videos (and optionally text prompts) as input conditions, making it impossible to specify sounds that users may want to combine with the generated audio.

Few studies have explored audio conditioning in video-to-audio synthesis to address their respective problem setting.
MultiFoley~\cite{chen2024video} uses conditional audio as a reference for the generated audio. 
Sketch2Sound~\cite{garcia2025sketch2sound} takes a similar approach, but only uses a particular set of signal features extracted from the original conditional audio to accept sonic or vocal imitations as conditions. 
Action2Sound~\cite{chen2024action2sound} focuses on disentangling foreground and ambient sound, using conditional audio to specify the appearance of ambient sound in the generated audio.
ReWaS~\cite{Jeong2025ReWaS} introduces audio-energy conditioning predicted from video into a text-to-audio generator via ControlNet \cite{Zhang2023ctrlnet}, enabling audio generation that is temporally aligned with the video.
Concurrent to our work, SelVA~\cite{lee2025selva} studies text-conditioned selective V2A via text-guided modulation of the video encoder.
Unlike these studies, we utilize audio conditioning to specify what kind of audio {\it should not} appear in the generated audio, enabling step-by-step generation in video-to-audio synthesis.

\subsection{Generative \emph{add} operation}

Generative \emph{add} operations in the audio domain are executed to generate audio that can be mixed with an input audio signal, often guided by a text prompt. 
These operations have been explored in text-to-audio~\cite{wang2023audit,jia2025audioeditor} and text-to-music~\cite{han2024instructme,parker2024stemgen,mariani2024multi,postolache2024generalized,karchkhadze2025simultaneous} synthesis and can be divided into two approaches: training-based and training-free.

In the training-based approach, the model is explicitly trained to perform the \emph{add} operation given the input audio. 
This training requires a triplet comprising an input audio, a text prompt, and an audio to be added as training data~\cite{wang2023audit,han2024instructme,parker2024stemgen}. 
Unfortunately, these methods are difficult to apply in our setting because such data is hard to obtain. 
Even within a single scene, a mixture of many sounds can be observed, and separating them into individual ones is challenging~\cite{owens2018audio,zhu2020visually,song2023visually}.
SonicVisionLM~\cite{Xie2024SonicVisionLM} introduces a timestamp-conditioned video-and-text-to-audio model that first converts video into text and then generates audio via a T2A backbone. While this design avoids audio contamination from visual features and enables additive video-to-audio generation, it loses the fine-grained audiovisual cues captured by multimodal V2A models~\cite{polyak2025moviegencastmedia,chen2024multifoley,cheng2024taming,Jun2025klingfoley}. As a result, it struggles with subtle or weakly visible events and relies on specialized video–audio–text–timestamp datasets that are costly to build.

On the other hand, the training-free approach is more flexible, as it leverages a pre-trained text-to-audio/music model without requiring any specific training. 
The \emph{add} operation is conducted as a partial generation of multi-track audio~\cite{mariani2024multi,postolache2024generalized,karchkhadze2025simultaneous} or a re-generation with a target prompt from structured noise obtained through inverting the input audio~\cite{jia2025audioeditor}. 
Instead of specific training data, these methods require particular properties in the pre-trained model: multi-track joint generation~\cite{mariani2024multi,karchkhadze2025simultaneous}, data-space diffusion models~\cite{postolache2024generalized}, and specific types of model architectures~\cite{jia2025audioeditor}, which limit their applicability to our video-to-audio setting.

Similar \emph{add} operations have been explored in computer vision, where models generate an object image to be added to an input background image. 
They are also categorized into training-based~\cite{singh2024smartmask,canberk2024erasedraw} and training-free approaches~\cite{tewel2025add}, but in either case, they rely on segmentation models to create training data or to guide the generation process. 
This means that a particular subset of pixels in the image is assumed to be entirely replaced by the generated pixels via the \emph{add} operation. 
As \emph{add} in the audio domain involves mixing rather than replacing, these approaches cannot be directly applied to our setting.

We adopted a training-based approach for this study, utilizing the \emph{add} operation in the audio domain, and designed our framework to eliminate the need for specialized training data. 
This approach makes it more practical and adaptable for video-to-audio synthesis tasks.

\section{Method}
\label{sec:method}

\subsection{Preliminaries}

\paragraph*{Generative modeling with flow-matching.}

Let $p_1(x)$ be a data distribution where $x \in \mathbb{R}^d$.
Flow matching~\cite{lipman2023flow} considers the probability flow ODE $\frac{\mathrm{d}}{\mathrm{d}t}\phi_t(x) = u_t(\phi_t(x))$, where $t \in [0, 1]$ is a timestep, $u_t$ is the velocity field, and $\phi_t(x) = \phi(x, t): \mathbb{R}^d \times \mathbb{R} \rightarrow \mathbb{R}^d$ is the flow that maps $x$ to the intermediate data $x_t$.
$\phi_t$ can be an arbitrary function that satisfies the terminal condition $\phi_1(x_1) = x_1$ and $\phi_0(x_1) \sim p_0$, where $p_0$ is a tractable distribution such as a standard normal distribution $\mathcal{N}(0, I)$.
Following the most popular setting, we define $\phi_t(x_1) = tx_1 + (1-t)x_0$, where $x_0 \sim  N(0, I)$, resulting in $u(\phi_t(x)) = x_1 - x_0$.

Solving the ODE from $t = 0$ to $1$ with an initial sample $x_0 \sim p_0(x)$ enables sampling from the target data distribution $x_1 \sim p_1(x)$.
To achieve this, a neural network is trained to predict $u_t (\phi_t(x))$, which corresponds to $x_1-x_0$ in our case, by minimizing the squared error over both data and timesteps.
In text-conditioned video-to-audio synthesis, the model takes additional conditional inputs, which are the input video $V$ and text prompt $C$, to model a conditional flow $u_t(\phi_t(x) | V, C)$.

\paragraph*{Guidance for flow-matching models with multiple conditions.}

Classifier-free guidance~\cite{ho2021classifier} is widely used to improve generation quality and fidelity to conditions. 
This guidance is typically conducted under a single condition, and it is nontrivial to extend it to two conditions, as in text-conditioned video-to-audio synthesis. 
To derive a proper guidance process, $p(x|V,C)$ is decomposed as follows:
\begin{align}
    p(x|V,C) = p(x) \brc*{\frac{p(x|V)}{p(x)}} \brc*{\frac{p(x|V,C)}{p(x|V)}}.
    \label{eq:cond_prob_v2a}
\end{align}
Given this decomposition,  Kushwaha and Tian~\cite{kushwaha2024vintage} proposed the following guided flow:
\begin{align}
    \tilde{u}_\theta(x_t) = u_\theta(x_t, t, \varnothing, \varnothing)
      &+ w_1 (u_\theta(x_t, t, V, \varnothing) - u_\theta(x_t, t, \varnothing, \varnothing)) \nonumber \\
      &+ w_2 (u_\theta(x_t, t, V, C) - u_\theta(x_t, t, V, \varnothing)), 
      \label{eq:guide_vintage}
\end{align}
where $\theta$ is a set of the model parameters, and $\varnothing$ denotes a null condition. 
The three terms of the right-hand side of Eq. (\ref{eq:guide_vintage}) respectively correspond to the three factors on the right-hand side of Eq. (\ref{eq:cond_prob_v2a}).
They empirically show that setting $w_1 = w_2$ achieves better results. 
In this case, $u_\theta(x_t, t, V, \varnothing)$ cancels out, which gives us the following simplified formulation:
\begin{align}
    \tilde{u}_\theta(x_t) = u_\theta(x_t, t, \varnothing,\varnothing) + w_1 (u_\theta(x_t, t, V, C) - u_\theta(x_t, t, \varnothing,\varnothing)).
      \label{eq:guide_vintage_simple}
\end{align}

\subsection{Problem setting: Step-by-step video-to-audio synthesis}

We are interested in iteratively generating complementary audio that complements previously generated audio.
Let $\xcond$ be the audio generated at the previous step, $\xtgt$ be the target audio to be generated at this step, $C_2$ be the text prompt that specifies the sound event for $\xtgt$ missed by $\xcond$, and $V$ be the input video.
Our goal is to sample $\xtgt$ from $p\brc*{\xtgt \middle| V, C_2, \xcond}$ so that $\xtgt$ corresponds to the concept described by $C_2$, semantically and temporally aligns with $V$, and does not contain duplicated audio present in $\xcond$.

From a straightforward standpoint, learning to generate samples from this distribution would require tuples of $(\xtgt, V, C_2, \xcond)$ as training data. 
Unfortunately, constructing such data from video is a challenging task known as visually-guided audio source separation~\cite{owens2018audio,zhu2020visually,song2023visually}, and thus, we cannot expect high-quality training datasets. 
Instead, we propose an alternative training framework that eliminates the need for such specialized data.

Note that this processing can be applied to the following generation step without loss of generality.
In the $k$-th generation step, we can set the mix of the previously generated $(k-1)$ audios as $\xcond$, and use it to sample $\xtgt$. 
Please refer to Section \ref{sec:evaluation_setup} for more details of this procedure.

\subsection{Formulation of target distribution with concept negation}

\begin{figure}[t]
    \centering
    \includegraphics[width=\linewidth]{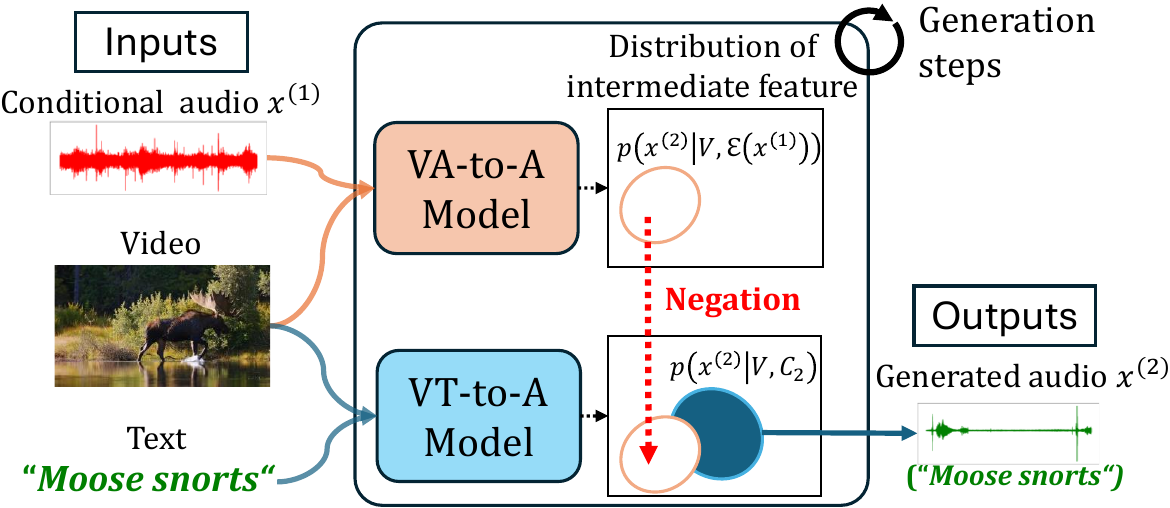}
    \caption{Overview of the proposed method. Each audio track should represent a distinct audio event. Using previously generated audio tracks as conditioning signals for the negation concept, we explicitly push the current generation process away from the already generated audio tracks.}
    \label{fig:method_overview}
\end{figure}

Recall that we want to generate $\xtgt$ to only cover a complementary sound event in $\xcond$. 
In this sense, conditioning by $\xcond$ corresponds to a concept negation~\cite{Du2020compositional,liu2022compositional,valle2025fugatto} in generating $\xtgt$; the generated $\xtgt$ {\it should not} contain any concepts related to $\xcond$.
We denote this type of audio condition as $\notcondc = \neg \condc$ to explicitly differentiate it from a standard type of audio condition denoted by $\condc$.
Based on the above-mentioned relationship between $\xcond$ and $\xtgt$, we approximate the target distribution of $\xtgt$ using the concept negation as follows:
\begin{align}
    p\brc*{\xtgt \middle| V, C_2, \xcond} \approx p\brc*{\xtgt \middle| V, C_2, \notxcondc}.
\end{align}
Following the study by Du et al.~\cite{Du2020compositional}, we assume that the concept negation holds
\begin{align}
    \label{eq:concept_neg}
    p\brc*{x, c_p, \neg c_n} \propto p(x) p(c_p | x) p(c_n | x)^{-1},
\end{align}
where $c_p$ and $c_n$ denote conditional concepts to generate $x$.

To derive our guidance, we decompose the target distribution using Eq. (\ref{eq:concept_neg}) and Bayes' theorem as:
\begin{align}
    p\brc*{\xtgt \middle| V, C_2, \notxcondc} \nonumber \\
    &\hspace{-6em}\propto p\brc*{\xtgt, V, C_2, \notxcondc} \nonumber \\
    &\hspace{-6em}\propto p\brc*{\xtgt, V} p\brc*{C_2 \middle| \xtgt, V} p\brc*{\xcondc \middle| \xtgt, V}^{-1} \nonumber \\
    &\hspace{-6em}\propto p\brc*{\xtgt} \!
            \brc*{\frac{p\brc*{\xtgt | V}}{p\brc*{\xtgt}}} \!
            \brc*{\frac{p\brc*{\xtgt \middle| V, C_2 }}{p\brc*{\xtgt \middle| V}}} \!
            \brc*{\frac{p\brc*{\xtgt \middle| V}}{p\brc*{\xtgt \middle| V, \xcondc}}}.
    \label{eq:2nd_posterior}
\end{align}
Similar to the derivation of Eq. (\ref{eq:guide_vintage}), we can derive the guidance process based on this decomposition, as shown in the next section. 
This indicates that we can sample $\xtgt$ using the new guidance with flow-matching models. 
Iterating this process enables step-by-step generation in video-to-audio synthesis.

\section{Implementation}
\label{sec:impl}

\subsection{Guided flow for step-by-step generation}

The decomposition shown in Eq. (\ref{eq:2nd_posterior}) yields a new guidance formulation comprising four terms: one unconditional flow term and three guidance terms. However, adjusting the coefficients of the three guidance terms is cumbersome in practice. 
Given the empirical results in VinTAGe~\cite{kushwaha2024vintage}, where simplifying the guidance by removing $u_\theta\brc*{x_t, t, V, \varnothing}$ in Eq. (\ref{eq:guide_vintage}) performs well, we also set the guidance coefficients so that $u_\theta\brc*{x_t, t, V, \varnothing}$ cancels out for simplification. Specifically, we set the sum of the coefficients of the first and third guidance terms to equal the coefficient of the second guidance term (see the Appendix for details). This leads to the following guided flow:
\begin{align}
     \label{eq:nag}
     \tilde{u}_{\theta,\psi}(x_t) = u_\theta(x_t, t, \varnothing, \varnothing)
      &+ \alpha \brc*{u_\theta(x_t, t, V, C_2) - u_\theta(x_t, t, \varnothing, \varnothing)} \nonumber \\
      &+ \beta \brc*{u_\theta\brc*{x_t, t, V, C_2} - u_{\theta,\psi}\brc*{x_t, t, V, \varnothing, \xcond}},
\end{align}
where $\alpha$ and $\beta$ are the coefficients of the guidance terms. As we require an audio-conditioned flow in the last term, we introduce an additional set of trainable parameters $\psi$ to adapt the text-conditioned video-to-audio model for this prediction, as detailed in the next subsection. 

The second term on the right-hand side of Eq. (\ref{eq:nag}) corresponds to a standard guidance term in text-conditioned video-to-audio models, which appeared in Eq. (\ref{eq:guide_vintage_simple}).
It strengthens the fidelity of the generated audio to the conditional video and text prompt.
The third term is a new guidance term introduced in our proposed method that pushes the generated audio away from the conditional audio $\xcond$. 
This prevents the already-generated audio events from being re-generated during the current generation step, enabling step-by-step generation without overlapping audio events. 
Since $\xcond$ is used similarly for a negative prompt, we refer to this new guidance as \PMfull~(\PMabbr).

\subsection{Training Flow estimator for \PMfull}

All the flows appearing on the right-hand side of Eq. (\ref{eq:nag}), except for the last one, can be estimated using standard text-conditional video-to-audio models. The remaining term is a conditional flow corresponding to the distribution $p(\xtgt|V, \xcondc)$. 
As $\condc$ is a standard type of audio conditioning, this flow estimator can be viewed as an extension of the video-to-audio model, incorporating conditional audio as an additional input. 
Therefore, we train the flow estimator using ControlNet~\cite{Zhang2023ctrlnet} parameterized by $\psi$.
This audio-conditioned flow estimator is trained to remain visually grounded while leveraging acoustic context from the reference audio, and is applied negatively at inference as a repulsive prior to avoid re-instantiating components already present in previous tracks.
As a base video-to-audio model, we used MMAudio, parameterized by $\theta$, for its high capability in video-to-audio synthesis.

\paragraph*{Model architecture.}
\label{sec:model_arch}

\begin{figure}[t]
    \centering
    \includegraphics[width=\linewidth]{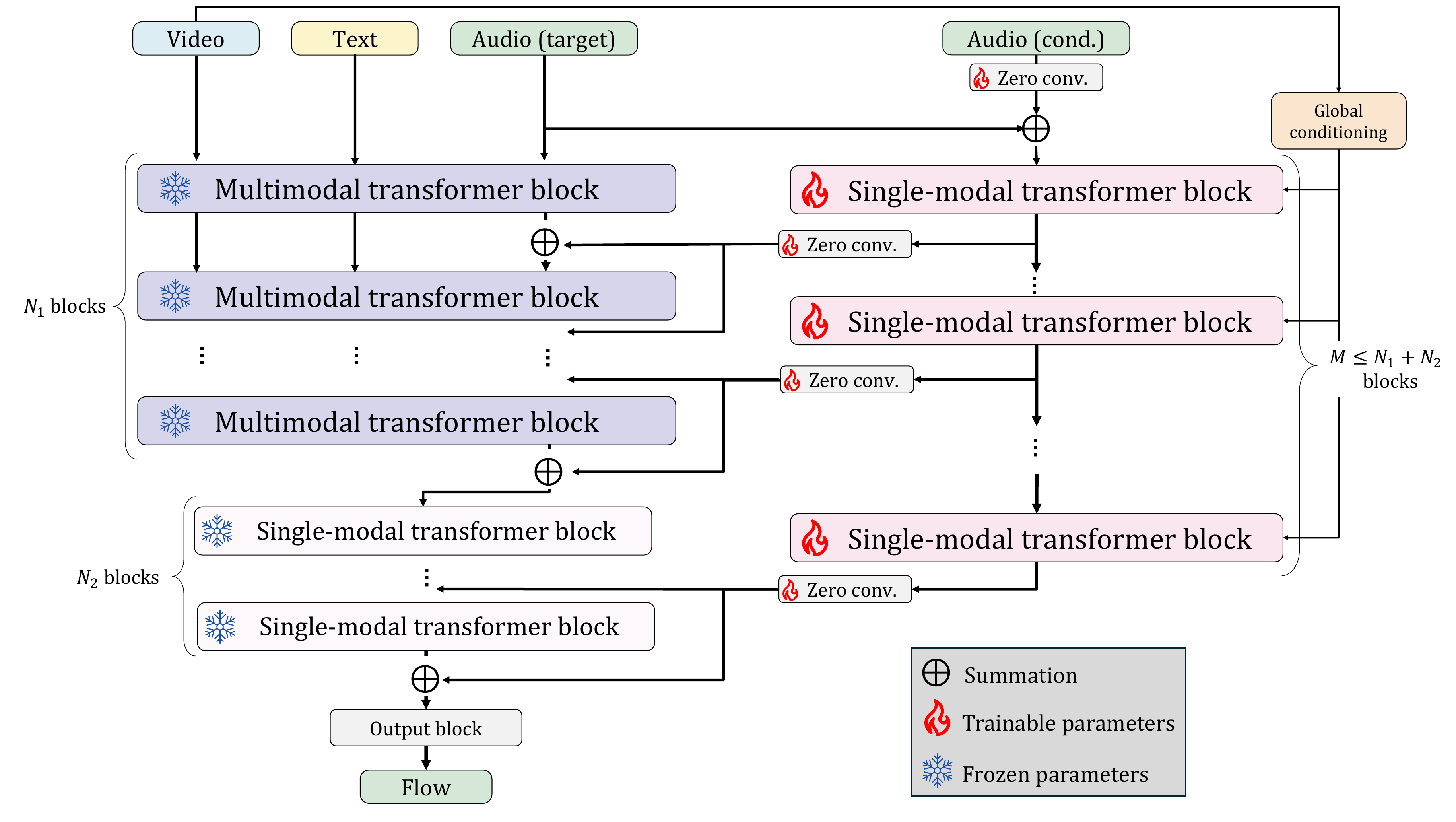}
    \caption{Overview of network architecture for the audio-conditional flow estimator. We adopt ControlNet for the multi-modal diffusion transformer (MM-DiT) to incorporate audio conditioning into the pre-trained MMAudio.}
    \label{fig:controlnet}
\end{figure}

Figure \ref{fig:controlnet} shows an overview of the ControlNet architecture of the flow estimator.
Since MMAudio uses a sophisticated architecture that extends MM-DiT~\cite{Esser2024sd3}, we adapt the ControlNet architecture accordingly. 
Inspired by Stable Diffusion 3.5~\cite{stabilityai2024sd35}, we stack several single-modal transformer blocks to extract features from the conditional audio, and the features extracted at each block are added to the intermediate features of the main branch in MMAudio.
During training, we freeze the pre-trained parameters of MMAudio and only update the parameters of the additional modules. 
Please refer to the Appendix for more details on model architecture, training, and inference.

\paragraph*{Training dataset.}
We follow the training strategy of MMAudio, jointly using text-video-audio and text-audio paired datasets. 
Specifically, we used VGGSound~\cite{chen2020vggsound} as a text-video-audio dataset, while Clotho~\cite{drossos2020clotho}, AudioCaps~\cite{kim2019audiocaps}, and WavCaps~\cite{mei2024wavcaps} were used as text-audio datasets. 
From each audio clip, we sampled a four-second audio segment as $x^{\mathrm{tgt}}$ and another one as $x^{\mathrm{cond}}$ so that the two segments do not overlap.
With this adjacent split, the two non-overlapping segments are more likely to share acoustic context (\eg, environment and recording conditions).
We also extracted a video segment corresponding to $x^{\mathrm{tgt}}$ as a conditional video $V$ when the data came from VGGSound; otherwise, we set the pretrained empty token of MMAudio as $V$. 
Then, the sampled clips were used to compute the flow $u_{\theta, \psi} \brc*{x^{\mathrm{tgt}}_t, t, V, \varnothing, x^{\mathrm{cond}}}$, and $\psi$ is optimized by minimizing the flow-matching loss.
Importantly, we use non-overlapping segments to encourage the audio-conditioned branch to extract higher-level acoustic cues (\eg, environment, timbre, recording conditions) that are shared across the video.

\paragraph{Design rationale for non-overlapping training pairs.}
Our goal at inference is to suppress re-generation of the \emph{mixture of previously generated tracks for the same time window}.
To instantiate \PMabbr, we introduce an audio-conditioned branch to predict $u_{\theta,\psi}(x_t, t, V, \varnothing, x^{\mathrm{cond}})$, which provides the flow direction induced by the conditioning audio.
In Eq.~\eqref{eq:nag}, this audio-conditioned estimator is applied with a negative sign.
This yields a repulsive guidance that steers the sampling trajectory away from making the sample more consistent with the conditioning audio.

During training, we optimize the audio-conditioned branch to follow the audio condition (positive guidance), \ie, to generate audio consistent with the acoustic cues in $x^{\mathrm{cond}}$ while remaining grounded in $V$.
We sample $(x^{\mathrm{tgt}}, x^{\mathrm{cond}})$ from non-overlapping segments of the same clip to encourage the branch to capture shared \emph{acoustic context} (e.g., environment, timbre, and recording conditions) that is more likely to be shared within a video.
In practice, this clip-level context provides a useful signal for repelling components already present in previous tracks at inference, while avoiding a degenerate solution that copies the conditioning waveform.
This is a good match for NAG because the branch is used \emph{negatively}: it only needs to identify directions that would increase consistency with $x^{\mathrm{cond}}$, which we then subtract during sampling.

\section{Experiments}
\label{sec:expr}

\subsection{Evaluation setup}

\label{sec:evaluation_setup}

\paragraph{Multi-Caps VGGSound: multi-captioned audio-video dataset for evaluation.}
We constructed a new audiovisual dataset, Multi-Caps VGGSound, to evaluate step-by-step video-to-audio generation.
We generated five captions using Qwen2.5-VL \cite{Qwen2.5-VL} for each video in the test split of the VGGSound dataset, totaling 15,221 video clips.
We instructed the model to produce captions that describe different sound events that could plausibly occur in the scene, including both foreground and background audio.
Since Qwen2.5-VL does not accept audio as input, captions are generated solely from visual inputs and may therefore deviate from the original audio (e.g., off-screen sounds).
However, this deviation is not an issue for our benchmark goal, which is to evaluate visually plausible and complementary sound authoring from video rather than reproducing the original recording.
Please refer to the Appendix for more details.

\paragraph{Task setup: Step-by-step audio generation.} 
We generated five audio tracks $\left\{x^{(k)} | k \in \{1, 2, \dots 5\}\right\}$ corresponding to audio captions $\left\{C_k | k \in \{1, 2, \dots 5\}\right\}$ for each video $V$ in the Multi-Caps VGGSound dataset.
The generation order is determined based on the semantic similarity between the video and the caption, so that the model generates core events first (See the Appendix for more details).
We extracted the first 8-second segment from the video and generated sounds for this segment using different captions.
Given the multiple generated audio tracks, we synthesized a composite audio $\tilde{x}$ by $\tilde{x} = \mathrm{normalize}(\Sigma_k{x^{(k)}})$.
We employed loudness normalization \cite{steinmetz2021pyloudnorm} as a simple mixing strategy for the composition, ensuring that the total loudness remained consistent with that of natural audio.
The target loudness was set to -20 LUFS, which corresponds to the mean loudness of the VGGSound test set.

\paragraph{Step-by-step audio generation with \PMabbr.}
To generate the audio tracks step-by-step using \PMabbr, we used the composite audio of all audio tracks generated in the previous generation steps as the conditioning signal for \PMabbr.
Specifically, at the generation step for $x^{(k)}$, we synthesized a composite audio  $\tilde{x}_{:k} = \mathrm{normalize}\brc*{\Sigma_l^{k-1}x^{(l)}}$ for the condition.
We generated the first audio only using the standard classifier-free guidance in Eq.~\eqref{eq:guide_vintage_simple}, since no audio track had been generated at the first generation step.
For the guidance coefficients, we empirically set $\alpha=4.5$ and $\beta=1.5$ in Eq.~\eqref{eq:nag} (see the Appendix for a sensitivity analysis).

\paragraph{Baseline models.}
We compared our proposed method with open-source text-and-video conditional audio (TV2A) generation models.
We chose each State-of-the-Art TV2A model among various training approaches: Seeing-and-Hearing~\cite{xing24seeing} as a TV2A model adapted from the T2A model without training, FoleyCrafter~\cite{zhang2024foleycrafter} as a TV2A model adapted from the T2A model through fine-tuning, and MMAudio~\cite{cheng2024taming} as a TV2A model trained from scratch.
Note that our model is built upon MMAudio with additional audio conditions introduced by the proposed ControlNet architecture.
We compared the proposed \PMabbr~to the original MMAudio-S-16k model with the classifier-free guidance (CFG) or negative prompting.

We tested two generation processes to generate multiple audio tracks and obtain a composite audio using the baseline models: independent generation and step-by-step generation based on negative prompts.
In the independent generation, we generated five sounds per video using different text conditions.
In this case, each generation process does not access the other audio tracks or their corresponding captions.
In the step-by-step generation based on negative prompts, we generated each audio track using negative prompting~\cite{woolf2024negativeprompting} to ensure it was distinct from all the captions used in the previous generation steps.
Specifically, for the video $V$ with the $k$-th audio caption $C_k$, we computed the guided flow by $\tilde{u}_\theta(x_t) = u_\theta(x_t, t, \varnothing,\varnothing) + w_1 \brc*{u_\theta(x_t, t, V, C_k) - u_\theta(x_t, t, \varnothing, C_{k, \mathrm{neg}})}$ at each timestep, where $C_{k,\mathrm{neg}}$ is the concatenation of the other captions $\left\{ C_{l} | l < k \right\}$.

\paragraph{Evaluation metrics.}
We assessed the quality of both the composite audio and the individual audio tracks to evaluate the step-by-step audio generation.

Following MMAudio~\cite{cheng2024taming}, we evaluated the composite audio in terms of audio quality, semantic alignment, and temporal alignment.
We assessed the audio quality of the generated audio using Fréchet Distance (FD), Kullback–Leibler (KL) distance, and Inception Score (IS) \cite{salimans2016improved}.
We used PANNs~\cite{kong2020panns} (\fdpanns) and VGGish~\cite{gemmeke2017audio} (\fdvgg) for computing FD, and PANNs (\klpanns) and PaSST (\klpasst) for computing KL, and PANNs for computing IS, respectively.
We assessed the semantic alignment between the input video and the composite audio by the cosine similarity between their embeddings extracted by ImageBind~\cite{girdhar2023imagebind} (IB-score).
We assessed temporal alignment between the input video and the generated audio using Synchformer \cite{iashin2024synchformer} (DeSync), where we took two 4.8-second segments at the beginning and end and averaged their scores.

For each audio track evaluation, we assessed its quality across four aspects: audio separability between audio tracks generated for the same video, audio quality, audio-text alignment, and audio-video alignment.
Since distinct audio components should be represented in separate audio tracks, each generated audio track should differ from the other tracks.
To evaluate audio separability, we computed the similarity between the CLAP~\cite{wu2023large} audio embeddings for each pair of audio tracks (10 pairs per video).
For Audio-Text alignment, we computed the similarity between CLAP text embeddings from the input prompts (used to generate each audio track) and the corresponding CLAP audio embeddings.
For audio quality and audio-video alignment, we adopted IS and IB-score, respectively, as in the composite audio evaluation protocol.

\subsection{Main results}
\begin{table*}[t]
\small
    \centering
    \caption{Quantitative evaluation of the composite audio synthesized from the generated multiple audio tracks. The results of one-step generation using a fused caption are shown as a reference.}
    \begin{NiceTabular}{l@{\hspace{4pt}}c@{\hspace{4pt}}c@{\hspace{4pt}}c@{\hspace{4pt}}c@{\hspace{1pt}}c@{\hspace{1pt}}c@{\hspace{1pt}}c}
    \toprule
    & \multicolumn{5}{c}{{\footnotesize Audio Quality}} & \multicolumn{2}{c}{\footnotesize A-V Align.} \\
    \cmidrule(lr{\dimexpr 4\tabcolsep-2pt}){2-6}
    \cmidrule(lr{\dimexpr 4\tabcolsep-2pt}){7-8}
    Method & {\scriptsize \fdpanns$\downarrow$} & {\scriptsize \fdvgg$\downarrow$} & {\scriptsize \klpanns$\downarrow$} & {\scriptsize \klpasst$\downarrow$} & {\scriptsize IS$\uparrow$} & {\scriptsize IB-score$\uparrow$} & {\scriptsize DeSync$\downarrow$} \\
    \midrule
    \multicolumn{8}{l}{\rowcolor{headerColor} {\textbf{One-Step Generation with Fused Caption (no separated audio track)}} \hfill} \\
    \deemph{Seeing-and-Hearing} & \deemph{25.42} & \deemph{5.68} & \deemph{2.81} & \deemph{2.76} & \deemph{6.45} & \deemph{36.88} & \deemph{1.22}  \\
    \deemph{FoleyCrafter} & \deemph{16.93} & \deemph{2.29} & \deemph{2.60}  & \deemph{2.52} & \deemph{11.93} & \deemph{27.78} & \deemph{1.23}  \\
    \deemph{MMAudio-S-16k} & \deemph{6.75} & \deemph{1.03} & \deemph{2.09} & \deemph{2.04} & \deemph{13.66} & \deemph{29.45} & \deemph{0.46}  \\
    \midrule
    \multicolumn{8}{l}{\rowcolor{headerColor} {\textbf{Independent Generation}} \hfill} \\
    Seeing-and-Hearing & 31.81  & 7.68 & 3.10 & 2.65 & 4.12 & 20.16 & 1.19  \\
    FoleyCrafter & 20.04 & 3.23 & 2.70 & 2.36 & 9.21 & 25.26 & 1.18  \\
    MMAudio-S-16k & 7.76 & 1.35 & 2.02 & 1.84 & 10.42 & 28.13 & \textbf{0.42} \\
    \midrule
    \multicolumn{8}{l}{\rowcolor{headerColor} {\textbf{Step-by-Step Generation with Negative Prompting}} \hfill} \\
    FoleyCrafter & 22.34 & 4.64 & 2.94 & 2.47 & 6.02 & 18.83 & 1.19  \\
    MMAudio-S-16k & 9.21 & 1.77 & 2.15 & 1.89 & 9.08 & 25.89 & 0.45 \\
    \midrule
    \multicolumn{8}{l}{\rowcolor{headerColor} {\textbf{Step-by-Step Generation with Negative Audio Guidance}} \hfill} \\
    \rowcolor{highlightColor} Ours & \textbf{6.47} & \textbf{0.98} & \textbf{2.01} & \textbf{1.76} & \textbf{10.58} & \textbf{28.65} & \textbf{0.42} \\
    \bottomrule
    \end{NiceTabular}
    \label{tab:composite_audio_evaluation}
\end{table*}
\begin{table*}[t]
\small
    \centering
    \caption{Quantitative evaluation of individual audio tracks. Our proposed method improves audio separability among multiple tracks while maintaining other scores.}
    \begin{NiceTabular}{l@{\hspace{4pt}}c@{\hspace{4pt}}c@{\hspace{4pt}}c@{\hspace{4pt}}c}
    \toprule
    & {\footnotesize Separability} & {\footnotesize Quality}  & {\footnotesize A-T Align.} &{\footnotesize A-V Align.} \\
    \cmidrule(lr{\dimexpr 4\tabcolsep-2pt}){2-2}
    \cmidrule(lr{\dimexpr 4\tabcolsep-2pt}){3-3}
    \cmidrule(lr{\dimexpr 4\tabcolsep-2pt}){4-4}
    \cmidrule(lr{\dimexpr 4\tabcolsep-2pt}){5-5}
    Method & \clapaa$\downarrow$ & IS$\uparrow$ & \clapta$\uparrow$ & IB-score$\uparrow$ \\
    \midrule
    MMAudio-S-16k & 79.75 & \textbf{12.47} & \underline{28.36} & \textbf{27.76} \\    
    MMAudio-S-16k + Neg. Prompting & \underline{75.57} & 11.19 & 27.14 & 24.53 \\
    \rowcolor{highlightColor} MMAudio-S-16k + NAG (Ours) & \textbf{71.38} & \underline{12.01} & \textbf{28.91} & \underline{26.67} \\
    \bottomrule
    \end{NiceTabular}
    \label{tab:soundtrack_evaluation}
\end{table*}

\paragraph{Objective evaluation on the composite audio.}
\Cref{tab:composite_audio_evaluation} shows the quantitative evaluation of the composite audio.
Our proposed method achieves the best results for all metrics except IS among all the methods.
We also evaluated the baseline models' one-step generation with the caption created by fusing the five captions for each video.
Though this generation process does not provide each audio track and differs from our goal of step-by-step generation, these values indicate the best possible scores of each baseline model.

\paragraph{Objective evaluation on each audio track.}
\Cref{tab:soundtrack_evaluation} shows the quantitative evaluation of the individual audio tracks.
The vanilla MMAudio-S-16k with CFG struggles to generate well-separated sounds for each audio track, as reflected in a lower audio separability score, although it achieves high audio quality and A-V Alignment.
Using negative prompting improves the audio separability but drastically degrades all the other scores.
Using \PMabbr\ also successfully improves the audio separability while maintaining high audio quality and A-V alignment.
The A-T alignment score marginally improves from that of the vanilla MMAudio.
We hypothesize that lower contamination by other audio concepts yields a higher A-T alignment score.

\begin{figure}[t]
    \centering
    \includegraphics[width=\linewidth]{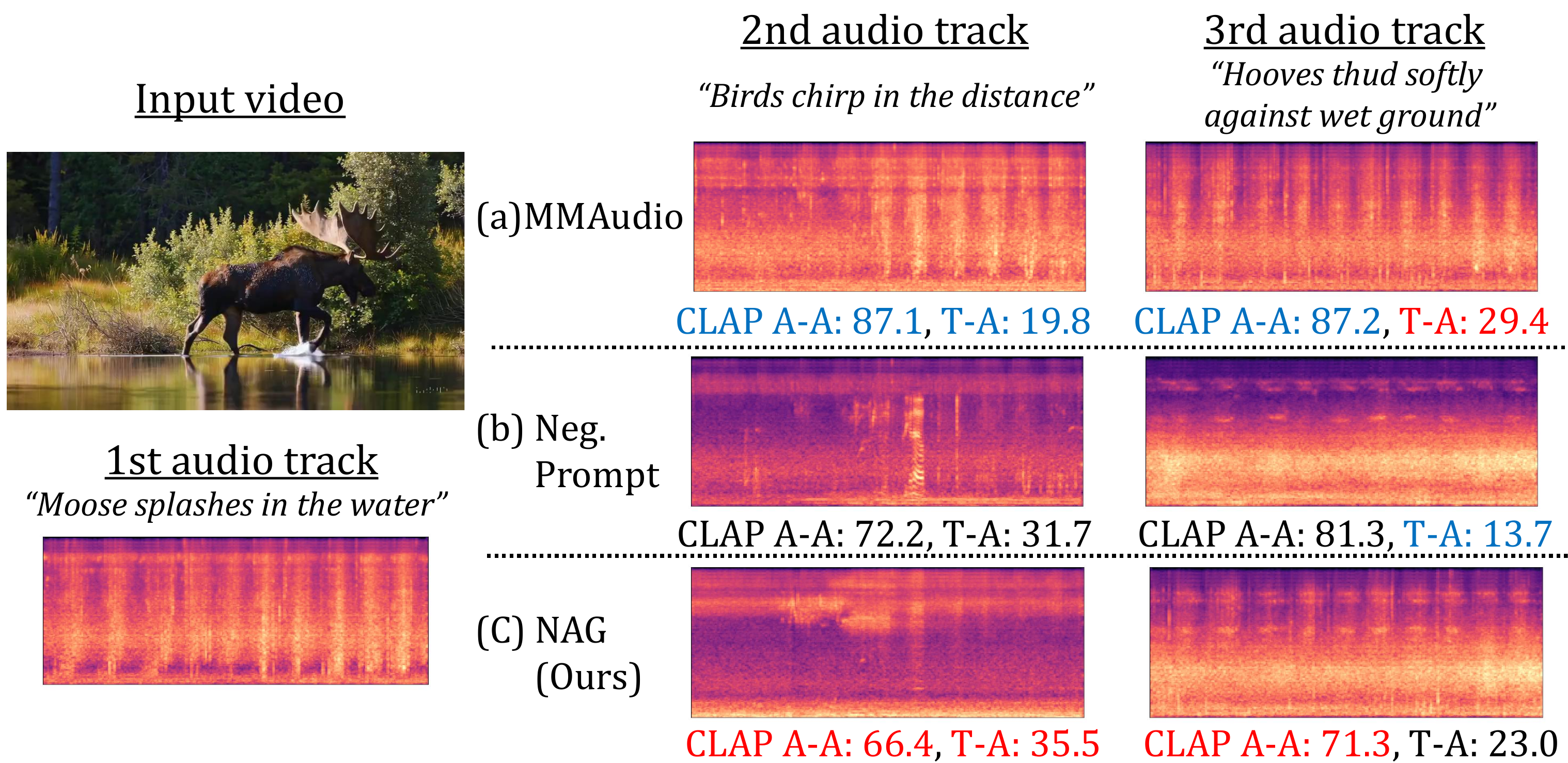}
    \caption{Spectrogram visualizations of step-by-step audio generation using (a) vanilla MMAudio, (b) MMAudio with negative prompting, and (c) MMAudio with Negative Audio Guidance (NAG). The best and worst CLAP A-A and T-A scores are highlighted in red and blue, respectively. The first audio track is generated using (a) in all settings, resulting in identical outputs. Our proposed method effectively suppresses previously generated sounds in subsequent steps (second and third tracks) while maintaining high alignment with the target text prompts.}
    \label{fig:step-by-ste-visualize}
\end{figure}

\paragraph{Visual comparison with baselines.}
\Cref{fig:step-by-ste-visualize} visually compares these three methods.
The first audio track is the same across all methods because it is generated using only CFG.
The vanilla MMAudio-S-16k tends to generate similar audio across multiple audio tracks.
All generated audio tracks from the vanilla MMAudio-S-16k contain the sound of water as the moose walks (visually shown as vertical segments that appear at regular intervals).
Since the moose and its movement are prominent in the input video, MMAudio tends to include the sound related to them regardless of the input text prompt.
This is also reflected in the higher IB-score, indicating that all audio tracks are semantically aligned with the input video, regardless of whether the input prompt represents background audio (as in the case of the second audio track).
Using negative prompting suppressed this contamination of the audio content, but it tends to suffer from worse text alignment.
In contrast, the proposed \PMabbr\ successfully suppressed the audio components already generated in the early generation steps while achieving better text alignment.
It generates a complementary sound specified by the text prompt by explicitly steering the generation process away from previously generated sounds. 
Please refer to the Appendix for additional generated samples.

\paragraph{Subjective evaluation.}
We also conducted a user study for subjective evaluation.
Table \ref{sup:tab:user_study_win_rate} presents the user study results for the final composite audio, showing that our method is preferred for audio quality, semantic alignment, and temporal alignment compared to the baseline.
\Cref{sup:tab:user_study_indv_avg} shows the results for the individual audio tracks.
Our method received significantly higher ratings for separability and marginally higher ratings for both audio quality and text fidelity.
Please refer to the Appendix for a more detailed statistical analysis and details of the user study setup.

\begin{table*}[t]
\small
    \centering
    \caption{The win rates of our proposed method (MMAudio-S-16k + NAG) against MMAudio-S-16k for composite audio. 95\% confidence intervals are reported as $\pm X$.}
    \begin{NiceTabular}{l@{\hspace{8pt}}c@{\hspace{8pt}}c@{\hspace{8pt}}c}
    \toprule
     & Audio quality$\uparrow$ & Semantic alignment$\uparrow$ & Temporal alignment $\uparrow$ \\
    \midrule
    Win rate & \textbf{71.36} \pmnum{7.71} & \textbf{76.00} \pmnum{7.62} & \textbf{61.14} \pmnum{7.85} \\
    \bottomrule
    \end{NiceTabular}
    \label{sup:tab:user_study_win_rate}
\end{table*}
\begin{table*}[t]
\small
    \centering
    \caption{Average ratings for individual audio tracks generated by the baseline and our method. 95\% confidence intervals are reported as $\pm X$.}
    \begin{NiceTabular}{l@{\hspace{8pt}}c@{\hspace{8pt}}c@{\hspace{8pt}}c}
    \toprule
    Method & Separability$\uparrow$ & Audio quality$\uparrow$ & Text fidelity $\uparrow$ \\
    \midrule
    MMAudio-S-16k & 2.24\pmnum{0.15} & 2.89\pmnum{0.14} & 2.42\pmnum{0.18}\\
    \rowcolor{highlightColor} MMAudio-S-16k + \PMabbr\ (Ours) & \textbf{3.35}\pmnum{0.15} & \textbf{3.30}\pmnum{0.14} & \textbf{3.12}\pmnum{0.18} \\
    \bottomrule
    \end{NiceTabular}
    \label{sup:tab:user_study_indv_avg}
\end{table*}

\begin{figure}[t]
    \centering
    \includegraphics[width=\linewidth]{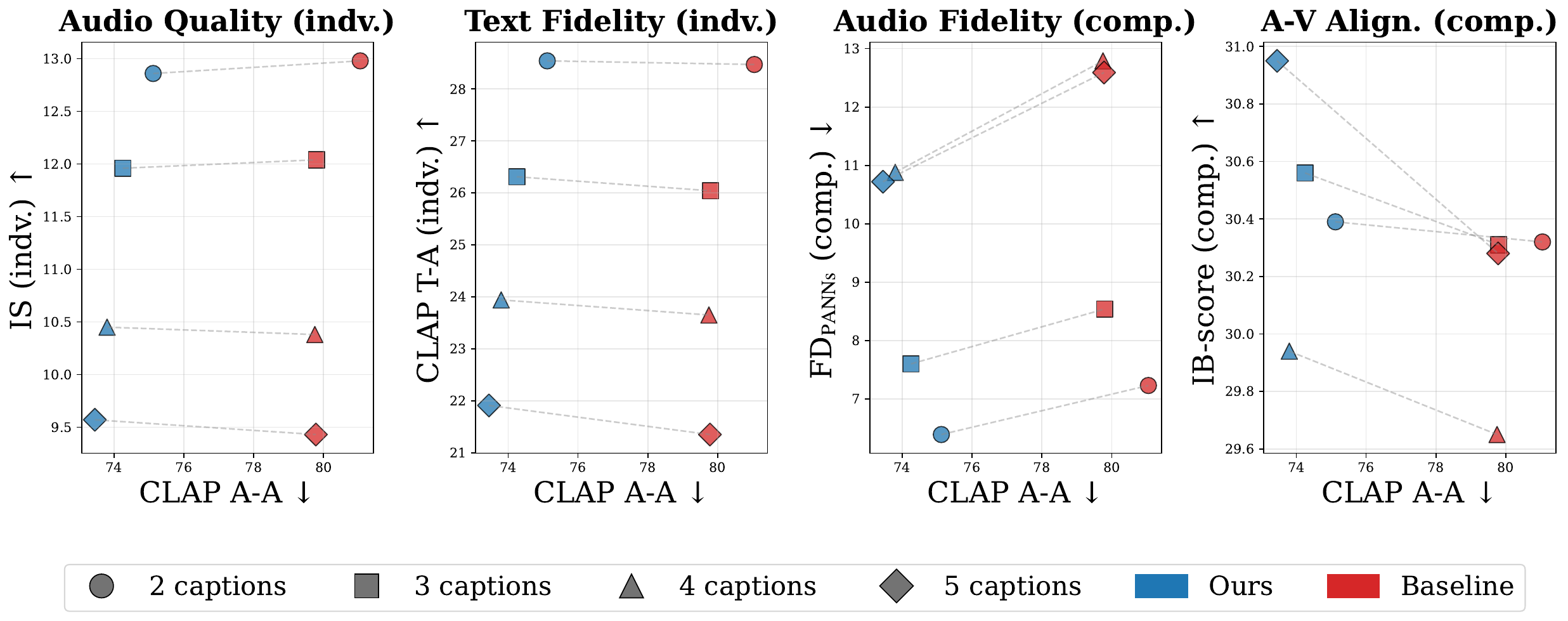}
    \caption{Analysis of the number of captions using VGGSounder. We compare MMAudio-S-16k (Baseline) and MMAudio-S-16k + NAG (Ours) on individual-track quality and text fidelity (left), and on composite-audio fidelity and audiovisual alignment (right). Audio separability is shown on the x-axis. Our method consistently improves separability without degrading quality, with larger gains as the number of captions increases.}
    \label{fig:sep_analysis}
\end{figure}

\subsection{Analysis on number of captions}
We used a fixed number of captions for Multi-Caps VGGSound to ensure consistent evaluation.
In real use cases, however, the number of sequential generation steps is chosen by the user based on the visual content and the desired number of audio events.
To examine whether our method remains effective across a broader range of plausible audio events, we conducted additional experiments on the VGGSounder~\cite{zverevwiedemer2025vggsounder} dataset.
VGGSounder provides a flexible number of human-annotated captions per video, reflecting both the visual content and the ground-truth audio.
We selected videos with 2–5 captions and evaluated the baseline model (MMAudio-S-16k) and our method (MMAudio-S-16k + NAG) on each subset separately.

\Cref{fig:sep_analysis} summarizes the results.
The two plots on the left report audio quality and text fidelity for the individual tracks, while the two plots on the right report audio fidelity and audiovisual alignment for the composite audio.
All plots use the audio separability of each track on the x-axis.
Across all caption counts, the baseline model exhibits consistently low separability, whereas our method reliably improves separability without sacrificing audio quality or text alignment.
The improvement becomes more pronounced as the number of captions increases.
For composite audio evaluation, our method consistently enhances both audio fidelity and audiovisual alignment.
Overall, these results demonstrate that the proposed step-by-step generation is beneficial even when only a few sound events are present, and its advantage increases as the number of audio events grows.

\section{Conclusion}
We introduced a novel video-to-audio generation method, guided by text, video, and audio conditions, that enables step-by-step synthesis.
By applying negative audio guidance alongside a text prompt, our approach generates multiple complementary sounds for the same video input, facilitating high-quality composite audio synthesis.
Importantly, our method does not require specialized training datasets.
We built it on a pre-trained video-to-audio model by adapting ControlNet for audio conditioning, enabling training on accessible single-reference datasets.
Quantitative and subjective evaluations show that our method improves the separability and text fidelity of generated audio at each step and the quality of the final composite audio.
\section*{Acknowledgment}
We sincerely thank Koichi Saito, Khaled Koutini, and Naoki Murata for their valuable comments and suggestions on an earlier version of this manuscript.


%
%
\bibliographystyle{splncs04}
\bibliography{reference}

\newpage

\onecolumn{ 
    \centering 
    \Large 
    \textbf{\paperTitle} \\ \vspace{0.5em}Supplementary Material \\ \vspace{1.0em} 
} 

\beginsupplement{}
\section{User study}
\label{sup:sec:user_study}
We conducted a user study to perform a subjective assessment on the Multi-Caps VGGSound dataset.
We used independent generation with MMAudio-S-16k as the baseline, corresponding to the case without negative audio guidance ($\beta=0$ in Eq. (9)), to assess the effectiveness of our proposed method.
We randomly sampled five video-caption sets from the dataset (each video has five captions) and generated five audio tracks for each video using the baseline and the proposed method.
As described in Section 5, we then synthesized composite audio for each video by mixing generated tracks, followed by loudness normalization.
In total, we showed 60 videos to each evaluator (50 videos with individual audio tracks and 10 videos with composite audio).
Human evaluators were asked to assess the quality of both individual audio tracks and the composite audio.

For the evaluation of individual audio tracks, evaluators rated each track on a scale from 1 to 5 (1-5; Poor, Subpar, Fair, Good, Excellent) across the following three aspects:
\begin{enumerate}
    \setlength{\itemsep}{0cm}
    \item  \textbf{Separability}: High if the audio does not contain content already present in previous audio tracks.
    \item \textbf{Audio quality}: High if the audio is free from noise, distortion, or artifacts.
    \item \textbf{Text fidelity}: High if the audio accurately reflects the caption.
\end{enumerate}
For the evaluation of composite audio, we performed A/B testing on pairs of composite audios, one generated by the baseline and the other by our method.
Specifically, five pairs of composite audios (each corresponding to the same video) were presented to evaluators, who were asked the following three questions for each pair:
\begin{enumerate}
    \setlength{\itemsep}{0cm}
    \item \textbf{Audio quality}: Which audio is of higher quality?
    \item \textbf{Semantic alignment}: Which audio has better semantic alignment with the video?
    \item \textbf{Temporal alignment}: Which audio has better temporal alignment with the video?
\end{enumerate}
For each question, evaluators could choose from three response options: "Audio A is better", "Audio B is better", and "Neutral".

\begin{figure}[t]
    \centering
    \includegraphics[width=\linewidth]{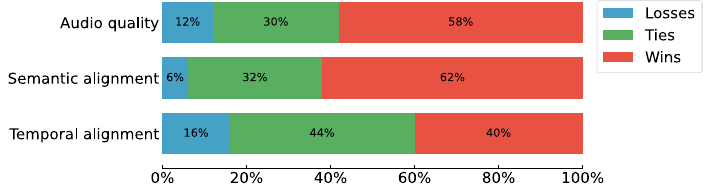}
    \caption{Results of user preference comparison between baseline (MMAudio-S-16k) and our method (MMAudio-S-16k with NAG) for composite audio. ``Win'' indicates the percentage of users who choose the composite audio generated by our method.}
    \label{fig:preference_composite}
\end{figure}

We collected 400 responses for the individual audio tracks (omitting an evaluation of the first track, as it is identical between the two methods) and 50 responses for the composite audio from 10 evaluators.
To verify statistical significance, we compute 95\% confidence intervals (CI) using the Wilson score interval for preference data and the standard error for rating scores. For the preference CI, ties are evenly split between wins and losses.
\Cref{fig:preference_composite} and Tables 3 and 4 in the main paper show the results.
Overall, the proposed method was preferred at least as much as the baseline across all evaluation criteria.

\section{Details of the formulation}
\label{sup:sec:proof_of_equations}

\subsection*{Proof of the Equation (5)}

Recall that the target distribution of each generation step can be written as:
\begin{align}
    p\brc*{\xtgt \middle| V, C_2, \notxcondc} &\propto p\brc*{\xtgt, V, C_2, \notxcondc} \nonumber \\
    &\propto p\brc*{\xtgt, V} \!
             \,p\brc*{C_2 \middle| \xtgt, V} \!
             \,p\brc*{\xcondc \middle| \xtgt, V}^{-1} \nonumber \\
    &= p\brc*{\xtgt} \!
       \,p\brc*{V \middle| \xtgt} \!
       \,p\brc*{C_2 \middle| \xtgt, V} \!
       \,p\brc*{\xcondc \middle| \xtgt, V}^{-1}.
    \label{sup:eq:2nd_posterior}
\end{align}

Using Bayes's theorem, we can decompose the last three terms in Eq. \eqref{sup:eq:2nd_posterior} as follows:

\begin{align}
    p\brc*{V \middle| \xtgt} &= \frac{p\brc*{\xtgt \middle| V}p\brc*{V}}{p\brc*{\xtgt}} \nonumber \\
    &\propto \frac{p\brc*{\xtgt \middle| V}}{p\brc*{\xtgt}},     \label{sup:eq:2nd_posterior_2nd_term} \\
    p\brc*{C_2 \middle| \xtgt, V} &= \frac{p\brc*{\xtgt \middle| V, C_2} p\brc*{C_2 \middle| V}}{p\brc*{\xtgt \middle| V}} \nonumber \\
    &\propto \frac{p\brc*{\xtgt \middle| V, C_2}}{p\brc*{\xtgt \middle| V}}, \label{sup:eq:2nd_posterior_3nd_term} \\
    p\brc*{\xcondc \middle| \xtgt, V} &= \frac{p\brc*{\xtgt \middle| \xcondc, V} p\brc*{\xcondc \middle| V}}{p\brc*{\xtgt \middle| V}} \nonumber \\
    &\propto \frac{p\brc*{\xtgt \middle| V, \xcondc}}{p\brc*{\xtgt \middle| V}}.
    \label{sup:eq:2nd_posterior_4th_term}
\end{align}

Note that we omit terms unrelated to $\xtgt$, since $\xtgt$ is the generation target.
Substituting Eqs. \eqref{sup:eq:2nd_posterior_2nd_term}, \eqref{sup:eq:2nd_posterior_3nd_term}, and \eqref{sup:eq:2nd_posterior_4th_term} into Eq. \eqref{sup:eq:2nd_posterior}, we get:

\begin{multline}
    p\brc*{\xtgt \middle| V, C_2, \notxcondc}  \\
    \propto p\brc*{\xtgt} \!
            \brc*{\frac{p\brc*{\xtgt | V}}{p\brc*{\xtgt}}} \!
            \brc*{\frac{p\brc*{\xtgt \middle| V, C_2 }}{p\brc*{\xtgt \middle| V}}} \!
            \brc*{\frac{p\brc*{\xtgt \middle| V}}{p\brc*{\xtgt \middle| V, \xcondc}}}.
    \label{sup:eq:2nd_posterior_identical}
\end{multline}

Therefore, Eq.~(5) holds.

\subsection*{Derivation of the Negative Audio Guidance in Equation (6)}

Similar to the guided flow proposed by Kushwaha and Tian~\cite{kushwaha2024vintage}, we can derive the guided flow corresponding to Eq.~(5) (or identically Eq. \eqref{sup:eq:2nd_posterior_identical}) as follows:

\begin{align}
     \tilde{u}_{\theta,\psi}(x_t) = u_\theta(x_t, t, \varnothing, \varnothing) \!
      &+ w_1' \brc*{u_\theta(x_t, t, V, \varnothing) - u_\theta(x_t, t, \varnothing, \varnothing)} \nonumber \\
      &+ w_2' \brc*{u_\theta(x_t, t, V, C_2) - u_\theta(x_t, t, V, \varnothing)} \nonumber \\
      &+ w_3' \brc*{u_\theta\brc*{x_t, t, V, \varnothing} - u_{\theta,\psi}\brc*{x_t, t, V, \varnothing, \xcond}},
      \label{sup:eq:nag_full}
\end{align}

where $w_1'$, $w_2'$, and $w_3'$ are the coefficients of the guidance terms.
The four terms on the right-hand side of Eq.~\eqref{sup:eq:nag_full} respectively correspond to the four factors on the right-hand side of Eq.~\eqref{sup:eq:2nd_posterior_identical}.
Following the empirical results provided by Kushwaha and Tian~\cite{kushwaha2024vintage}, we consider canceling out $u_\theta(x_t, t, V, \varnothing)$ for simplification. 
Specifically, we set $w_1' = \alpha$, $w_3' = \beta$, and $w_2' = \alpha + \beta$ as follows:
\begin{alignat}{1}
    \tilde{u}_{\theta,\psi}(x_t) \!
     = u_\theta(x_t, t, \varnothing, \varnothing) \nonumber
        &+ \alpha \Bigl(\cancel{u_\theta(x_t, t, V, \varnothing)} - u_\theta(x_t, t, \varnothing, \varnothing)\Bigr) \nonumber \\
        &+ (\alpha + \beta) \Bigl(u_\theta(x_t, t, V, C_2) - \cancel{u_\theta(x_t, t, V, \varnothing)}\Bigr) \nonumber \\
        &+ \beta \brc*{\cancel{u_\theta\brc*{x_t, t, V, \varnothing}} - 
        u_{\theta,\psi}\brc*{x_t, t, V, \varnothing, \xcond}}, 
\end{alignat}

which yields Eq.~(6).

\section{Details of model architecture, training, and inference}
\label{sup:sec:training_details}

We added one transformer block in the ControlNet for every two blocks in the main network (i.e., $N_1 + N_2 = 12$ and $M = 6$ in Fig.~3).
We set the channel dimensions and number of heads for multi-head attention to match the settings of MMAudio-S-16k. 
Our ControlNet has a total of 107M parameters, and the generation at each step, computed by this ControlNet using NAG, takes 2.07 seconds on an H100.

We followed the training setup of MMAudio~\cite{cheng2024taming} for training the ControlNet.
We used the AdamW optimizer with a learning rate of $10^{-4}$, $\beta_1 = 0.9$, $\beta_2 = 0.95$, and a weight decay of $10^{-6}$.
The network is trained for 200K iterations with a batch size of 512.
Compared to MMAudio's default of 300K iterations, we reduced the number of training steps to 200K, as we observed earlier convergence in our experiments.
We only updated the parameters of the ControlNet while fixing the pre-trained parameters of MMAudio, enabling more efficient training.
For learning rate scheduling, we applied a linear warm-up over the first 1K steps up to $10^{-4}$, followed by two reductions, each by a factor of 10, after 80\% and 90\% of the total training steps.
We used mixed-precision training with {\tt bf16}  for the training efficiency and trained on 8 H100 GPUs.
The entire training process, including evaluation on the validation and test sets every 20K iterations, took approximately 10 hours.
After training, we applied post-hoc EMA~\cite{Karras2024edm2} with a relative width $\sigma_{\mathrm{rel}} = 0.05$ to obtain the final parameters of the ControlNet.
\section{Details of the Multi-Caps VGGSound dataset}
\label{sup:sec:dataset_construction}
\begin{figure}[ht]
    \centering
    \includegraphics[width=\linewidth]{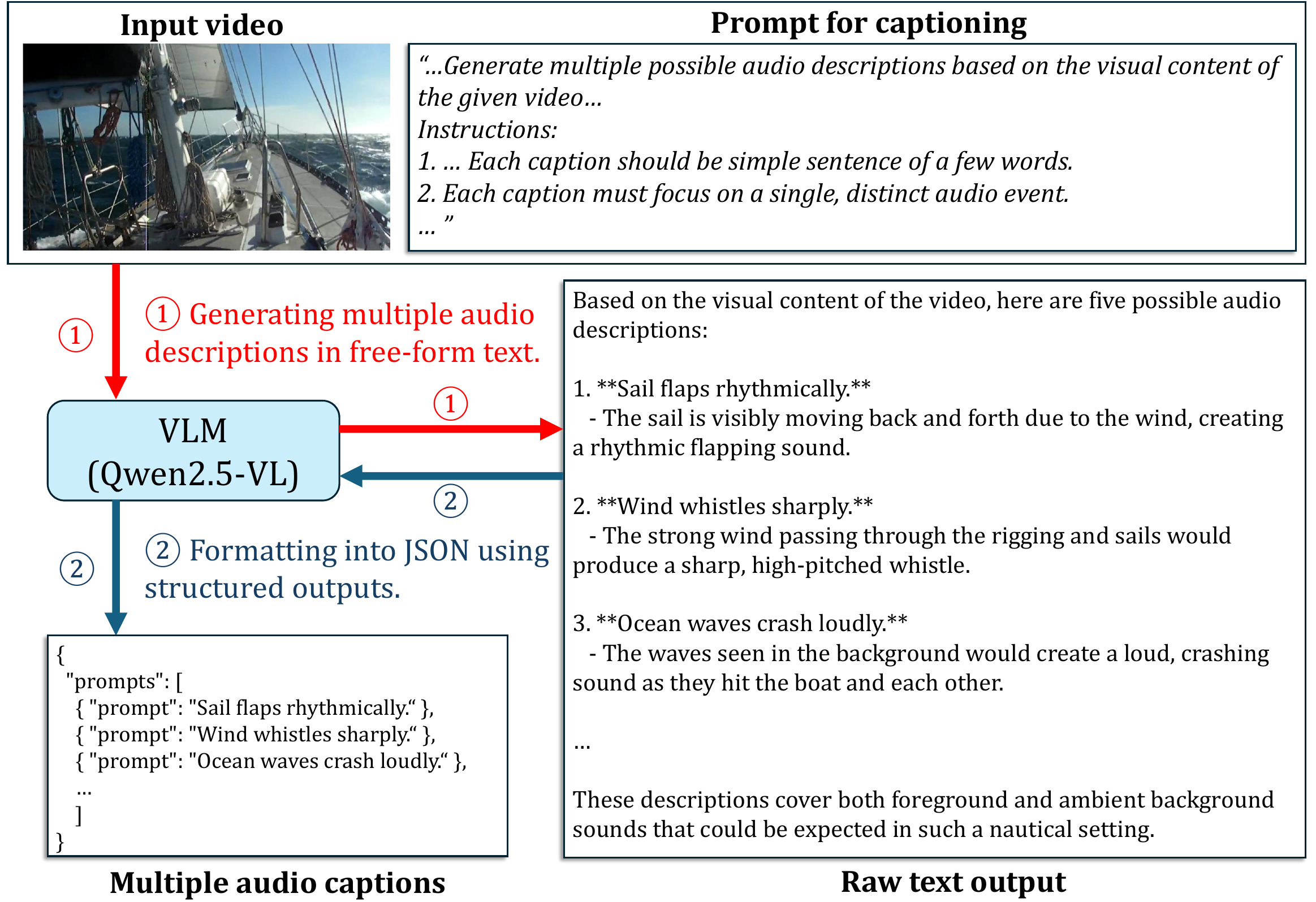}
    \caption{Overview of the dataset construction pipeline. Multiple audio captions were generated for each video using Qwen2.5-VL via a two-step process: free-form captioning followed by structured JSON formatting. The input prompt on the top is simplified; see Fig.~\ref{sup:fig:prompt} for the full version.}
    \label{sup:fig:dataset_construction}
\end{figure}

As described in Section 5.1, we generated five captions for each video in the test split of the VGGSound dataset using Qwen2.5-VL \cite{Qwen2.5-VL}, resulting in 76,105 video-caption pairs (15,221 videos $\times$ 5 captions).
\Cref{sup:fig:dataset_construction} shows an overview of the dataset construction workflow.

\begin{figure}[ht]
    \centering
    \includegraphics[width=\linewidth]{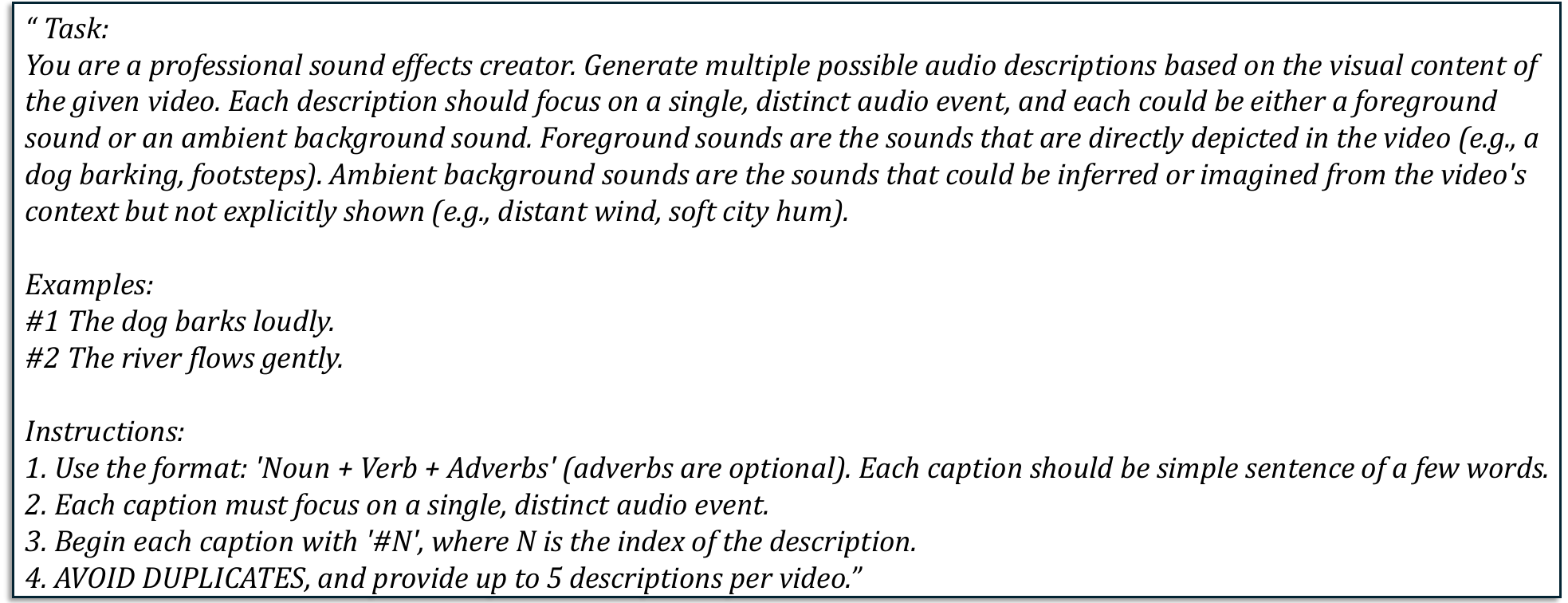}
    \caption{Full prompt for generating multiple possible audio captions.}
    \label{sup:fig:prompt}
\end{figure}

\begin{figure}[ht]
    \centering
    \includegraphics[width=\linewidth]{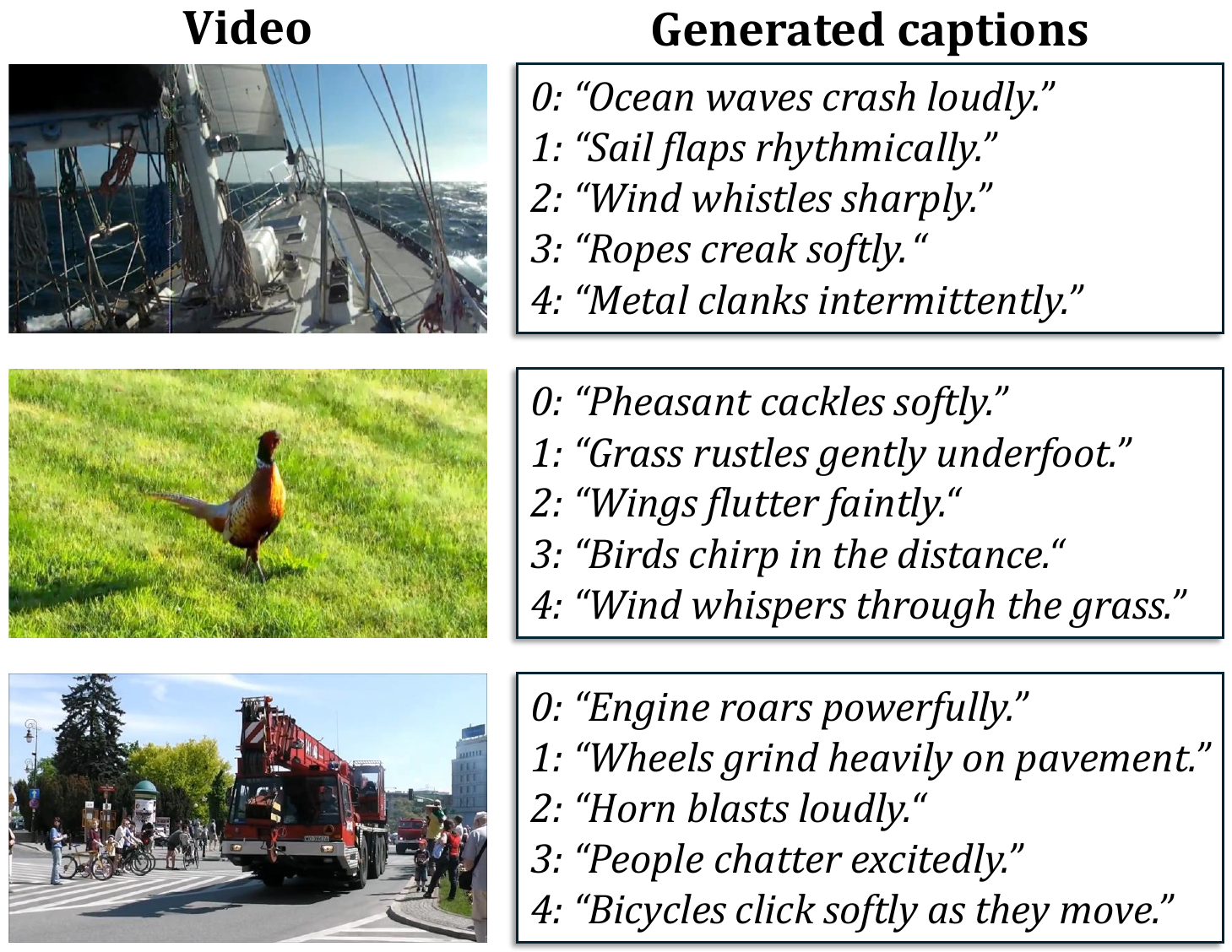}
    \caption{Examples of the Multi-Caps VGGSound dataset. We added multiple captions to the test split of the VGGSound dataset using Qwen2.5-VL, shown in \Cref{sup:fig:dataset_construction}.}
    \label{sup:fig:dataset_examples}
\end{figure}

We adopted a two-step approach to ensure that captions follow a unified format across all videos (i.e., short, simple sentences that describe distinct audio events).
Specifically, given an input video, we first generated multiple possible free-form audio captions describing the audio events likely present in the video.
Next, we reformatted the output into a structured JSON format using the generation of structured outputs (as implemented in vLLM~\cite{kwon2023efficient}).
The full prompt we used in the first step is shown in Fig.~\ref{sup:fig:prompt}.
While Qwen2.5-VL generates multiple captions in response to this prompt, the output may vary between inferences.
To standardize this, the second step converts the results into a unified JSON format that lists only the captions.
This structured format is well-suited for use as text conditions in text-conditional video-to-audio models.
Examples of video and caption pairs are shown in \Cref{sup:fig:dataset_examples}.

Using VLM-generated captions may introduce some noise, as not all descriptions perfectly match the actual audio.
However, we manually inspected a subset and confirmed that the captions remain visually plausible.
Since Foley workflows typically begin with a silent video and incrementally add plausible sounds based solely on visual context, this setup naturally aligns with Foley-style use cases in which the original recording does not match the target audio.

\section{Performance analysis across types of audio event}
To examine whether the proposed method improves separability consistently across audio event types, we first analyzed the label distribution of Multi-Caps VGGSound. 
We used GPT-5~\cite{OpenAI2025GPT-5} to categorize each caption into eight coarse VGGSound categories (Animals, Music, People, Vehicle, Home, Tools, Sports, and Nature) and 309 fine-grained labels. 
The coarse category distribution is as follows: Animals 23.8\%, Music 21.1\%, People 14.2\%, Vehicle 9.3\%, Home 9.1\%, Tools 8.8\%, Sports 7.4\%, and Nature 6.3\%. 
The fine-grained labels are nearly balanced, with 210--250 captions assigned to each label.

We then computed the CLAP A-A score for each fine-grained label to evaluate how separability improves across different types of audio events. 
Our method consistently improves the score for all 309 labels, with gains ranging from 2.25 to 9.64 points. 
The largest improvements are observed for temporally distinctive events, such as \textit{woodpecker pecking}, \textit{heartbeat}, \textit{bird calls}, and \textit{speedboat acceleration}. 
In contrast, the least-improved classes are mainly \textit{music, instrumental sounds}, or \textit{human vocal sounds}, which are often visually salient and therefore already strongly reinforced by the visual condition.

These results indicate that the overall improvement is not driven only by dominant categories and does not hide severe degradation in minority classes. 
Instead, the proposed method consistently improves audio separability across a wide range of event types, with gains especially pronounced for temporally distinctive sounds.
\section{Sensitivity analysis of the \PMfull\ coefficients}
\label{sup:sec:coefficients_sensitivity}
We conducted a sensitivity study on the guidance coefficients of \PMabbr\ ($\alpha$, $\beta$ in Eq.~(6).
Specifically, we varied $\alpha \in \{3.5, 4.5\}$ and $\beta \in \{0.0, 1.0, 1.5, 2.0\}$ and generated audio tracks for all combinations of these parameter pairs in \textbf{random generation order}.
The individual audio tracks and their corresponding composite audio were evaluated using the same setup and metrics described in Section 5.1.
We used \clapaa\ for audio separability, IS and \fdpanns\ for audio quality, \clapta\ for text fidelity, IB-score for semantic alignment with video, and DeSync for temporal alignment with video.

The results are summarized in \Cref{sup:tab:cfg_nag_sensitivity}.
Both \clapaa\ and \clapta\ for the individual audio tracks consistently improved with increasing $\beta$.
This indicates that \PMabbr\ effectively generates well-separated audio tracks with enhanced text alignment, likely due to reduced contamination from other audio concepts.
The best \fdpanns\ and IB-score are achieved at $\beta = 1.0$, indicating that a moderate strength of \PMabbr\ also enhances audio fidelity and semantic alignment with video.
Using a small $\alpha$ slightly deteriorates performance across all metrics, potentially due to the degradation of the performance of the base MMAudio (the default CFG strength recommended by Cheng et al.~\cite{cheng2024taming} is 4.5).
Considering trade-offs among these metrics, we selected $\alpha=4.5$ and $\beta=1.5$ as our default setting. 

\begin{table*}[t]
\small
    \centering
    \caption{Sensitivity study on the guidance coefficients of \PMabbr\ ($\alpha, \beta$ in Eq. (9)). We determined generation order by the random order strategy (see Section~\ref{sup:sec:generation_order} and \Cref{sup:tab:generation_order_comparison}) and the same order is applied across all settings. The first three metrics are computed on the individual audio tracks, and the last four are computed on the composite audio.  $\ast$: identical to the independent generation of MMAudio-S-16k with the default CFG strength of 4.5. $\dagger$: our default setting.}
    \begin{NiceTabular}{l@{\hspace{2pt}}c@{\hspace{2pt}}c@{\hspace{2pt}}c@{\hspace{4pt}}c@{\hspace{2pt}}c@{\hspace{2pt}}c@{\hspace{2pt}}c}
    \toprule
    & \multicolumn{3}{c}{{\footnotesize Individual Audio Tracks}} & \multicolumn{4}{c}{\footnotesize Composite Audio}  \\
    \cmidrule(lr{\dimexpr 4\tabcolsep-2pt}){2-4}
    \cmidrule(lr{\dimexpr 4\tabcolsep-2pt}){5-8}
    Setting & \clapaa$\downarrow$ & IS$\uparrow$ & \clapta$\uparrow$ & \fdpanns$\downarrow$ & IS$\uparrow$ & IB-score$\uparrow$ & DeSync$\downarrow$ \\
    \midrule
    $\alpha=3.5$, $\beta=0.0$ & 79.46 & 12.27 & 28.34 & 7.83 & 10.37 & 27.80 & 0.45 \\
    $\alpha=3.5$, $\beta=1.0$ & 75.22 & 12.06 & 28.75 & 7.52 & 10.17 & 27.94 & 0.45 \\
    $\alpha=3.5$, $\beta=1.5$ & 73.45 & 11.84 & 28.89 & 7.62 & 9.90 & 27.62 & 0.45 \\
    $\alpha=3.5$, $\beta=2.0$ & 71.97 & 11.62 & 29.01 & 7.84 & 9.69 & 27.28 & 0.45 \\
    \midrule
    $\alpha=4.5$, $\beta=0.0$$^\ast$ & 79.75 & 12.47 & 28.36 & 7.76 & 10.42 & 28.13 & 0.43 \\
    $\alpha=4.5$, $\beta=1.0$ & 76.21 & 12.22 & 28.69 & 7.30 & 10.38 & 28.38 & 0.43 \\
    \rowcolor{highlightColor} $\alpha=4.5$, $\beta=1.5$$^\dagger$& 74.74 & 12.00 & 28.79 & 7.32 & 10.21 & 28.15 & 0.43 \\
    $\alpha=4.5$, $\beta=2.0$ & 73.44 & 11.80 & 28.88 & 7.52 & 9.92 & 27.88 & 0.43 \\
    \bottomrule
    \end{NiceTabular}
    \label{sup:tab:cfg_nag_sensitivity}
\end{table*}
\section{One-step generation with post-processing}

To obtain multiple separated audio tracks, one possible approach is a one-step generation with post-processing: generating a composite audio from a fused caption, followed by track decomposition using audio source separation.
However, this approach fails for two reasons: (i) \textbf{one-step generation frequently misses audio events} when the text prompt contains multiple concepts, and (ii) \textbf{audio source separation degrades quality}, producing artifacts and unstable separated outputs.
We compare this approach with our proposed method to clarify its limitations.

We first generate single-track outputs using MMAudio-S-16k with fused captions (the composite sound results are shown in Table 1 in the main paper).
Then, we apply AudioSep~\cite{liu22w_interspeech,liu2023separate} to obtain separated audio tracks.
We use the official AudioSep model with its default configuration\footnote{https://github.com/Audio-AGI/AudioSep} and the same captions for separation as in our step-by-step approach.

\Cref{sup:tab:one_step_audiosep} shows a comparison between this approach and our method.
The separated tracks exhibit substantially worse text-audio and video-audio alignment (CLAP T-A and IB-Score) because the one-step results often fail to capture all required sound events.
These missing events lead to silent or noisy separated tracks, hurting both audio separability (CLAP A-A) and overall quality.
These results indicate that one-step generation with post-processing is a suboptimal strategy for producing multiple audio tracks, whereas our sequential method effectively mitigates these issues.
It is also worth noting that this approach complicates user interaction: revising only a specific sound event requires re-running both the generator and a separation model, which is unintuitive and fragile.
In contrast, our method provides an intuitive user interface that lets a user focus on a specific audio event at a time.

\begin{table*}[t]
\small
    \centering
    \caption{Comparison between one-step generation with audio source separation and our method.}
    \begin{NiceTabular}{l@{\hspace{2pt}}c@{\hspace{2pt}}c@{\hspace{2pt}}c@{\hspace{2pt}}c}
    \toprule
    & {\footnotesize Separability} & {\footnotesize Quality}  & {\footnotesize A-T Align.} &{\footnotesize A-V Align.} \\
    \cmidrule(lr{\dimexpr 4\tabcolsep-2pt}){2-2}
    \cmidrule(lr{\dimexpr 4\tabcolsep-2pt}){3-3}
    \cmidrule(lr{\dimexpr 4\tabcolsep-2pt}){4-4}
    \cmidrule(lr{\dimexpr 4\tabcolsep-2pt}){5-5}
    Method & \clapaa$\downarrow$ & IS$\uparrow$ & \clapta$\uparrow$ & IB-score$\uparrow$ \\
    \midrule
    MMAudio (one-step) $\rightarrow$ AudioSep & 79.11 & 9.04 & 23.42 & 21.96 \\    
    \rowcolor{highlightColor} MMAudio + NAG (Ours, step-by-step) & \textbf{71.38} & \textbf{12.01} & \textbf{28.91} & \textbf{26.67} \\
    \bottomrule
    \end{NiceTabular}
    \label{sup:tab:one_step_audiosep}
\end{table*}

\section{Performance Comparison with a Text-to-Audio Generator}

We compare our method with independent text-to-audio (T2A) generation using AudioX~\cite{tian2026audiox} on the Multi-Caps VGGSound dataset.
\Cref{sup:tab:comparison_with_t2a} summarizes the results.
AudioX achieves higher text alignment and audio quality than our method, but shows substantially lower video-audio alignment.
This result suggests that applying a T2A model to step-by-step V2A generation is insufficient for video synchronization, even when the prompts are generated from the video content.

In contrast, our method benefits from video conditioning and achieves better audio-video alignment. 
At the same time, the lower text-alignment score suggests that current V2A models still have room for improvement in prompt-level controllability. 
Stronger text alignment in V2A models could complement our step-by-step generation framework by enabling more precise control over individual tracks while maintaining synchronization with the video.

\begin{table}[t]
    \small
    \centering
    \caption{Comparison with independent T2A generation.1}
    \begin{NiceTabular}{l@{\hspace{5pt}}c@{\hspace{5pt}}c@{\hspace{5pt}}c@{\hspace{5pt}}c}
    \toprule
        & \multicolumn{2}{c}{{\footnotesize Individual Audio Tracks}} & \multicolumn{2}{c}{{\footnotesize Composite Audio}} \\
        \cmidrule(lr{\dimexpr 4\tabcolsep-2pt}){2-3}
        \cmidrule(lr{\dimexpr 4\tabcolsep-2pt}){4-5}
        {\footnotesize Method} & {\footnotesize CLAP T-A$\uparrow$} & {\footnotesize IS$\uparrow$} & {\footnotesize IB-score$\uparrow$} & {\footnotesize DeSync$\downarrow$} \\
        \midrule
        AudioX (T2A) & \textbf{38.90} & \textbf{12.30} & 19.54 & 1.25 \\
        \rowcolor{highlightColor} Ours & 28.91 & 10.58 & \textbf{28.65} & \textbf{0.42}  \\ 
    \bottomrule
    \end{NiceTabular}
    \label{sup:tab:comparison_with_t2a}
\end{table}
\section{Comparison of generation order}
\label{sup:sec:generation_order}

To study the effect of generation order, we ranked captions based on text-video similarity using the ImageBind score. 
We tested three variants: random order, descending order (where the core event is first), and ascending order (where the subtle event is first). 
The results are shown in Table~\ref{sup:tab:generation_order_comparison}.
Descending order provides the best results for all metrics, indicating that generating the prominent event in the video first is vital to improve the generation quality in step-by-step generation.

\begin{table*}[t]
\small
    \centering
    \caption{Comparison of generation order. We rank the captions based on the text-video ImageBind score and sort them in ascending and descending order.}
    \begin{NiceTabular}{l@{\hspace{4pt}}c@{\hspace{4pt}}c@{\hspace{4pt}}c@{\hspace{4pt}}c@{\hspace{1pt}}c@{\hspace{1pt}}c@{\hspace{1pt}}c}
    \toprule
    \multirow{2}{*}{Generation order} & \multicolumn{5}{c}{{\footnotesize Audio Quality}} & \multicolumn{2}{c}{\footnotesize A-V Align.} \\
    \cmidrule(lr{\dimexpr 4\tabcolsep-2pt}){2-6}
    \cmidrule(lr{\dimexpr 4\tabcolsep-2pt}){7-8}
    & \fdpanns$\downarrow$ & \fdvgg$\downarrow$ & \klpanns$\downarrow$ & \klpasst$\downarrow$ & IS$\uparrow$ & IB-score$\uparrow$ & DeSync$\downarrow$ \\
    \midrule
    MMAudio-S-16k & 7.76 & 1.35 & 2.02 & 1.84 & 10.42 & 28.13 & \textbf{0.42} \\
    \midrule
    Random & 7.32 & 1.24 & 2.02 & 1.78 & 10.21 & 28.15 & 0.43 \\
    Ascending & 7.36 & 1.26 & 2.02 & 1.78 & 10.02 & 28.10 & 0.43 \\    
    \rowcolor{highlightColor} Descending & \textbf{6.47} & \textbf{0.98} & \textbf{2.01} & \textbf{1.76} & \textbf{10.58} & \textbf{28.65} & \textbf{0.42} \\
    \bottomrule
    \end{NiceTabular}
    \label{sup:tab:generation_order_comparison}
\end{table*}
\section{Evaluation on AudioCaps and Movie Gen Audio Bench datasets}

To validate cross-dataset generalizability, we additionally evaluate our method on AudioCaps \cite{kim2019audiocaps} and Movie Gen Audio Bench \cite{polyak2024movie}.
We follow the same caption-generation procedure as Multi-Caps VGGSound using Qwen2.5-VL to generate multiple captions (see Section~\ref{sup:sec:dataset_construction}).
The results are shown in Tables~\ref{sup:tab:audiocaps_eval} and \ref{sup:tab:movie_gen_eval}.
Our method consistently improves audio separability and overall performance, demonstrating its generalizability across datasets.

\begin{table*}[t]
\small
    \centering
    \caption{Quantitative evaluation of both the composite audio and individual audio tracks generated from step-by-step generation on the AudioCaps test set.}
    \begin{NiceTabular}{l@{\hspace{4pt}}c@{\hspace{4pt}}c@{\hspace{4pt}}c@{\hspace{4pt}}c@{\hspace{4pt}}c@{\hspace{4pt}}c@{\hspace{4pt}}c}
    \toprule
    & \multicolumn{4}{c}{{\footnotesize Individual Audio Tracks}} & \multicolumn{3}{c}{{\footnotesize Composite Audio}} \\
    \cmidrule(lr{\dimexpr 4\tabcolsep-2pt}){2-5}
    \cmidrule(lr{\dimexpr 4\tabcolsep-2pt}){6-8}
    Method & CLAP A-A$\downarrow$ & IS$\uparrow$ & CLAP T-A$\uparrow$ & IB-Score$\uparrow$ & IS$\uparrow$ & IB-score$\uparrow$ & DeSync$\downarrow$ \\
    \midrule
    MMAudio-S-16k & 68.89 & \textbf{9.65} & 29.09 & \textbf{24.91} & 6.81 & 25.73 & 0.56\\
    \rowcolor{highlightColor}  + NAG (Ours) & \textbf{60.43} & 9.29 & \textbf{29.25} & 23.46 & \textbf{7.08} & \textbf{26.36} & \textbf{0.54} \\
    \bottomrule
    \end{NiceTabular}
    \label{sup:tab:audiocaps_eval}
\end{table*}
\begin{table*}[t]
\small
    \centering
    \caption{Quantitative evaluation of both the composite audio and individual audio tracks generated from step-by-step generation on the Movie Gen Audio Bench.}
    \begin{NiceTabular}{l@{\hspace{4pt}}c@{\hspace{4pt}}c@{\hspace{4pt}}c@{\hspace{4pt}}c@{\hspace{4pt}}c@{\hspace{4pt}}c@{\hspace{4pt}}c}
    \toprule
    & \multicolumn{4}{c}{{\footnotesize Individual Audio Tracks}} & \multicolumn{3}{c}{{\footnotesize Composite Audio}} \\
    \cmidrule(lr{\dimexpr 4\tabcolsep-2pt}){2-5}
    \cmidrule(lr{\dimexpr 4\tabcolsep-2pt}){6-8}
    Method & CLAP A-A$\downarrow$ & IS$\uparrow$ & CLAP T-A$\uparrow$ & IB-score$\uparrow$ & IS$\uparrow$ & IB-score$\uparrow$ & DeSync$\downarrow$ \\
    \midrule
    MMAudio-S-16k & 72.25 & 8.56 & 28.37 & \textbf{19.35} & 6.06 & 19.58 & 0.77\\
    \rowcolor{highlightColor}  + NAG (Ours) & \textbf{66.32} & \textbf{8.69} & \textbf{28.78} & 18.51 & \textbf{6.63} & \textbf{19.96} & \textbf{0.73} \\
    \bottomrule
    \end{NiceTabular}
    \label{sup:tab:movie_gen_eval}
\end{table*}
\section{Applicability beyond MMAudio} 
We additionally applied NAG to ControlFoley~\cite{yang2026controlfoleyunifiedcontrollablevideotoaudio}, which natively supports conditioning on text, video, and audio. 
The results in \ref{sup:tab:controlfoley} show that NAG consistently improves separability with par quality.

\begin{table}[h]
    \small
    \centering
    \caption{Generalization to ControlFoley. CLAP A-A/T-A: individual tracks; others: composite sound.}
    \begin{NiceTabular}{l@{\hspace{1pt}}c@{\hspace{4pt}}c@{\hspace{5pt}}c@{\hspace{4pt}}c@{\hspace{4pt}}c}
        \toprule
        & \multicolumn{2}{c}{{\footnotesize Individual Audio Tracks}} & \multicolumn{3}{c}{{\footnotesize Composite Audio}} \\
        \cmidrule(lr{\dimexpr 4\tabcolsep-2pt}){2-3}
        \cmidrule(lr{\dimexpr 4\tabcolsep-2pt}){4-6}
        {\footnotesize Method} & {\footnotesize CLAP A-A$\downarrow$} & {\footnotesize CLAP T-A$\uparrow$} & {\footnotesize IS$\uparrow$} & {\footnotesize IB-score$\uparrow$} & {\footnotesize DeSync$\downarrow$} \\
        \midrule
        {\footnotesize ControlFoley} & 60.67 & 31.45 & 10.21 & 28.18 & \textbf{0.391} \\
        \rowcolor{highlightColor} {\footnotesize + NAG} & \textbf{53.81} & \textbf{32.17} & \textbf{10.39} & \textbf{28.32} & \textbf{0.391} \\ 
        \bottomrule
    \end{NiceTabular}
    \label{sup:tab:controlfoley}
\end{table}
\section{Additional visualizations}
\begin{figure}[t]
    \centering
    \includegraphics[width=\linewidth]{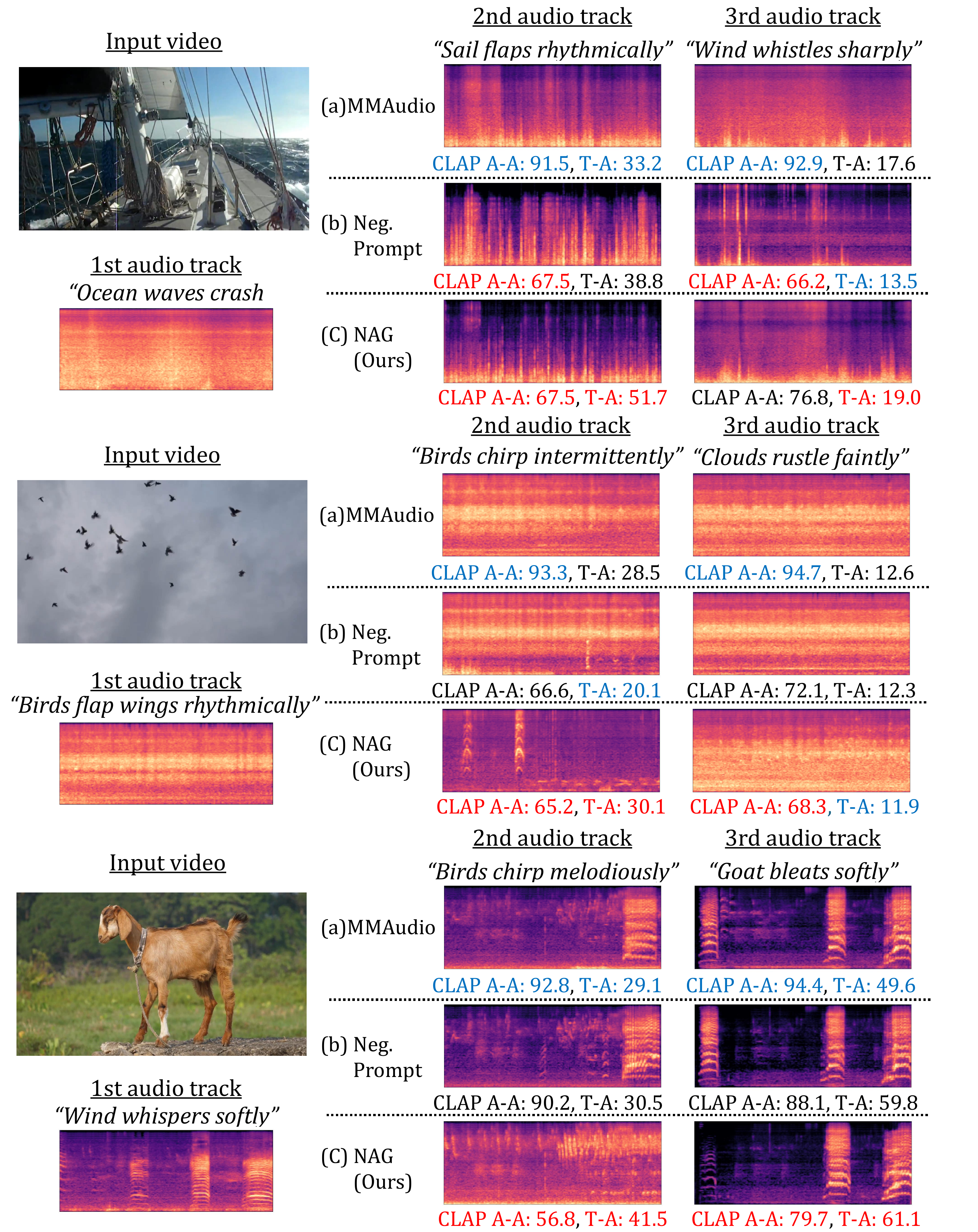}
    \caption{Additional spectrogram visualization. Our proposed method effectively suppresses previously generated sounds in subsequent steps while maintaining high alignment with the text prompts. The best and worst \clapaa\ and T-A scores are highlighted in red and blue, respectively.}
    \label{sup:fig:step-by-step-visualize}
\end{figure}

\label{sup:sec:additional_visualizations}
\Cref{sup:fig:step-by-step-visualize} shows additional spectrogram visualizations comparing MMAudio, MMAudio + negative prompting, and MMAudio + \PMabbr.
More generated samples are available at our project page: \url{https://ahykw.github.io/sbsv2a/}.

\section{Limitation}

\paragraph{Slight degradation of the audio quality of each audio track.}
In terms of individual audio track quality, our method marginally improves text fidelity but slightly degrades audio quality. 
While \PMabbr\ effectively eliminates contamination from other audio tracks, the outputs sometimes exhibit poor alignment with the text captions or suffer from low quality, such as silence or muffled sound.
This may stem from limitations in the base MMAudio model, particularly with handling subtle or rare sounds (e.g., “Carpet rustles gently”, “Wings flap gently”, “Snowflakes fall silently”, “Crowd murmurs quietly”). 
Even when conditioned only on such text prompts (text-to-audio generation), MMAudio often produces hums, noise, or unnaturally loud sounds, likely due to the scarcity of such audio events in its training data.
These mismatches suggest a domain gap between Multi-Caps VGGSound and MMAudio’s training distribution. 
Since \PMabbr\ only guides generation away from previous outputs, overall quality and text alignment rely heavily on the base TV2A model's capabilities.
The effectiveness of our proposed method would likely be more pronounced if the base model supported a broader range of text prompts (i.e., ideally broad enough to match the range supported by LLMs) and could generate more diverse audio outputs, even for the same video input.

\paragraph{Suboptimal audio mixing process.}
In this work, we synthesized composite audio using a simple mixing strategy, summing multiple audio tracks without weighting them, followed by loudness normalization.
While effective, it does not account for the natural loudness of each audio track and might be suboptimal. 
As optimal mixing can differ for each video and audio content, incorporating a generative model to support this process could further enhance the quality of the composite audio.
We leave this direction for future work.

\end{document}